\documentclass{article}

\usepackage[preprint]{neurips_2021}
\usepackage[utf8]{inputenc} % allow utf-8 input
\usepackage[T1]{fontenc}    % use 8-bit T1 fonts
\usepackage{hyperref}       % hyperlinks
\usepackage{url}            % simple URL typesetting
\usepackage{booktabs}       % professional-quality tables
\usepackage{amsfonts}       % blackboard math symbols
\usepackage{nicefrac}       % compact symbols for 1/2, etc.
\usepackage{microtype}      % microtypography
\usepackage{xcolor}         % colors

\usepackage{multirow}
\usepackage{dirtytalk}
\usepackage{mathtools}
\usepackage{natbib}
\setcitestyle{numbers,square}
\usepackage{dsfont}

\DeclareMathOperator*{\argmin}{arg\,min}

\title{GINA: \\ Neural Relational Inference From Independent Snapshots}

\author{%
  Gerrit Großmann\thanks{Corresponding author} \\
  Department of Computer Science\\
  Saarland Informatics Campus\\
  Saarland University\\
  Germany, 66123 Saarbrücken \\
  \texttt{gerrit.grossmann@uni-saarland.de} \\
   \And
  Julian Zimmerlin \\
  Department of Computer Science\\
  Saarland Informatics Campus\\
  Saarland University\\
  Germany, 66123 Saarbrücken \\
  \texttt{s8jjzimm@stud.uni-saarland.de} \\
   \And
  Michael Backenköhler \\
  Department of Computer Science\\
  Saarland Informatics Campus\\
  Saarland University\\
  Germany, 66123 Saarbrücken \\
  \texttt{michael.backenkoehler@uni-saarland.de} \\
   \And
  Verena Wolf \\
  Department of Computer Science\\
  Saarland Informatics Campus\\
  Saarland University\\
  Germany, 66123 Saarbrücken \\
  \texttt{verena.wolf@uni-saarland.de} \\
}

\begin{document}

\maketitle

\begin{abstract} %arise from local dynamis
Dynamical systems in which local interactions among agents give rise to complex emerging phenomena are ubiquitous in nature and society.
This work explores the problem of inferring the unknown interaction structure (represented as a graph) of such a system from measurements of its constituent agents or individual components (represented as nodes). 
We consider a setting where the underlying dynamical model is unknown and where different measurements (i.e., \emph{snapshots}) may be independent (e.g., may stem from different experiments).
We propose \texttt{GINA} (\texttt{G}raph \texttt{I}nference \texttt{N}etwork \texttt{A}rchitecture), a graph neural network (GNN) to simultaneously learn the latent interaction graph and, conditioned on the interaction graph, the prediction of a node's observable state based on adjacent vertices. 
\texttt{GINA} is based on the hypothesis that the ground truth interaction graph---among all other potential graphs---allows to predict the state of a node, given the states of its neighbors, with the highest accuracy. 
We test this hypothesis and demonstrate \texttt{GINA}'s effectiveness on a wide range of interaction graphs and dynamical processes.
%
%Our work is based on the hypothesis that the ground truth interaction graph allows it to predict a node's observable state---given the states of its neighbors---with the highest accuracy. 
%The interaction graph is represented as the first layer in the neural network and imposes a bottleneck of information flow using a powerful yet efficient counting abstraction. 
%The subsequent layers learn node-state probabilities. The bottleneck provides a natural regularization keeping the network from learning unnecessary edges. 
%, including models of epidemic spreading, neural spiking, opinion diffusion, evolutionary dynamics, and self-organized synchronization.
\end{abstract}

%%%%%%%%%%%%%%%%%%%%%%%%%%%%%%
%%%%%%%%%%%%%%%%%%%%%%%%%%%%%%
%  Introduction
%%%%%%%%%%%%%%%%%%%%%%%%%%%%%%
%%%%%%%%%%%%%%%%%%%%%%%%%%%%%%
\section{Introduction}
Complex interacting systems are ubiquitous in nature and society.
% da geht es ja noch nciht um network inference
%However, deciphering these interactions and inferring underlying structural constraints remains a major challenge.
However, their analysis remains challenging.
% This is partly due to the reductionist tradition in science, which sees a system as the sum of its parts.
% models? 
Traditionally, the analysis of complex systems is based on models of individual components.
This reductionist perspective reaches its limitations when the interactions of the individual components---not the components themselves---become the dominant force behind a system's dynamical evolution.
Understanding a system's functional organization   given such data is relevant for the analysis \cite{fornito2015connectomics,prakash2012spotting,amini2016resilience,finn2019use}, design \cite{zitnik2018modeling,hagberg2008rewiring,memmesheimer2006designing}, control \cite{gu2015controllability,grossmann2020learning}, and prediction \cite{kipf2018neural,zhang2019general} of many complex phenomena that are constrained by a graph topology or contact network.
%In recent years, driven by high throughput experiments and advances in model architectures, deep learning made remarkable progress in complex system analysis \cite{pathak2018model,kutz2017deep,mrowca2018flexible}.

Here, we focus on the internal interaction structure (i.e., graph or network) of a complex system.
We propose a  machine learning approach to infer this structure based on observational data of the individual components or constituent agents (i.e., nodes).
% This work proposes a graph machine learning approach to infer the internal interaction structure (i.e., graph or network) of a complex system based on observational data from the individual components or constituent agents (i.e., nodes).
% We call these observations \emph{snapshots}.
%We refer to these observations as \emph{snapshots} and assume that, for a snapshot, we simultaneously measure the states of all components at a specific point in time. 
We refer to these observations as \emph{snapshots} and assume that the observable states of all components are measured simultaneously.
However, we make no prior assumption about the relationship between snapshots.
Specifically, measurements may be taken from different experiments with varying initial conditions and at arbitrary time points.

%This differentiates our work from the rich pool of network reconstruction literature which is based on the analysis of time series data that have been proposed recently.

Most recent work focuses on time series data, where observations are time-correlated and
the interaction graph is inferred from the joint time evolution of the node-states   \cite{zhang2019general,kipf2018neural}.
Naturally, time series data typically contains more information on the system's interaction than snapshot data.
However, in many cases, such data is not available.
For instance, in some cases, one has to destroy a system to access their components (e.g., slice a brain \cite{rossini2019methods}, observe a quantum system \cite{martinez2019pileup}, or terminate a cell \cite{chan2017gene}).
Sometimes, the relevant time scale of the system is too small (e.g., in particle physics) or too large (e.g., in evolutionary dynamics) to be observed.
Often, there is a trade-off between spatial and temporal resolution of a measurement \cite{sarraf2016advances}.
In addition, measurements may be temporally decoupled due to large observation intervals and thus become unsuitable for methods that exploit correlations in time.
%Lastly, one might be interested in performing graph inference from historical data and only unstructured records from certain time points are preserved.
Yet,  machine learning techniques for graph inference from independent data remain underexplored in the literature.

\begin{figure}[t]
\centering
\includegraphics[width=0.98\textwidth ]{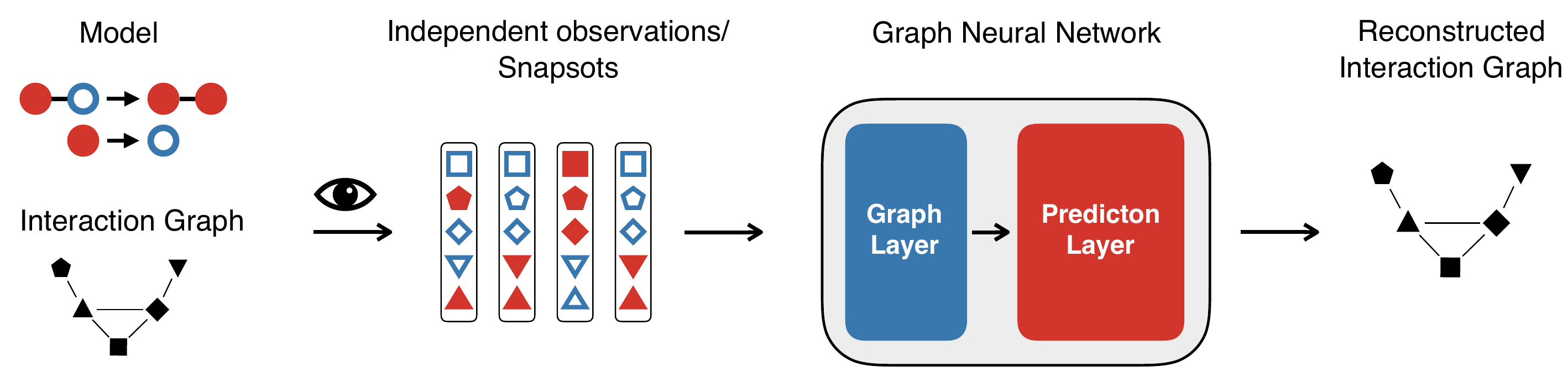}
\caption{Schematic illustration the problem setting. We are interested in inferring the latent interaction graph from observational data of the process. }
 \label{fig:overview}
\end{figure}

% welche Rolle spielt die Dynamik bzw. die Regeln nach denen die temporale Dynamik/Interaktion zw Agenten abläuft?

% Ein ganzer Absatz zu dem, was wir nicht tun...

% ACHTUNG, wurde zu related work geschoben!
%In our work, we assume no prior knowledge about the laws that govern the system dynamics, setting our approach apart from model-based reconstruction methods. 
%(e.g., methods specifically tailored for gene-network \cite{kishan2019gne,omranian2016gene} or protein-protein interaction network \cite{hashemifar2018predicting} reconstruction).
% ACHTUNG, wurde zu related work geschoben!
%Moreover, in contrast to most statistical methods, we aim at finding the interaction topology, even in the presence of non-linear dynamical laws governing the system.
%In such cases, one has to not only consider pair-wise correlations among components, but also the joint impact of all neighboring components.  

%Das stimmt ja so eigentlich nicht
%In this work, we do not make any assumptions about the nature of the temporal interactions. 
%We only assume that the observed patterns given by the node-states are generated by interaction rules that take into account the underlying network structure. 

% In this work, we aim as making as few assumptions as possible about the nature of the temporal interactions, setting our approach apart from the rich pool of model-based reconstruction methods.
In contrast to many state-of-the-art approaches, our approach is model-free without any assumptions on the dynamical laws.
Conceptually, our analysis is founded on identifying patterns in snapshots 
%that are characteristic of the process' dynamics to infer the interaction graph.
% Neu
We exploit that identified patterns are likely the result of probabilistic local interactions and, thus, carry information about the underlying connectivity.
% sind eigentlich beides keine guten beispiele
%In particular, this includes interaction patterns arising from stochastic and non-linear dynamics such as patterns of epidemic spreading or opinion formation. 
%Lastly, 

We propose an approach based on ideas recently popularized for time series based network inference \cite{zhang2018reconstructing,kipf2018neural}.
It provides an elegant way of formalizing the \emph{graph inference problem} with minimal parametric assumptions on the underlying dynamical model.
The core assumption is that the interaction graph \say{best describing} the observed data is the ground truth.

In the time series setting, this would be the graph that best enables time series forecasting.
% We assume the network that \say{best describes} the observed data is the ground truth.
Instead, we aim at finding the graph that best enables us to predict a node's state based on its direct neighborhood within a snapshot
and not at future times. 
To this end, we use a prediction model to learn a node's observable state (in the sequel: simply \emph{state} or \emph{node-state}) given the joint state of all adjacent nodes.
Then we maximize the prediction accuracy by jointly optimizing the interaction graph and the prediction model.
Loosely speaking, we assume that the information shared between a node and its complete neighborhood is higher in the ground truth graph than in other potential interaction typologies. 

%However, there is a catch to this approach.
However, in a trivial implementation, the network that enables the best node-state prediction is the complete graph because it provides all information available in a given snapshot. 
%The prediction model can then, if sufficiently powerful, learn corresponding weights and types of influences. 
This necessitates a form of regularization in which edges that are present, but not necessary, reduce prediction quality. We do this using a neighborhood aggregation scheme that acts as a bottleneck of information flow. 
%Specifically, we identify each node with a node-state (from a finite set of node-states).
We count the number of neighbors of a node $v$ in each node-state and use these counts to predict state probabilities of $v$. 
Thus, the prediction model represents the conditional probability of a node-state given an aggregation of the complete neighborhood.
Hence, irrelevant neighbors reduce the information content of the counting abstraction and consequently the prediction quality.
This neighborhood aggregating provides a powerful description for many relevant dynamical models of interacting systems \cite{gleeson2011high,fennell2019multistate}.

The node-states can directly relate to the measurements or be the result of an embedding or a discretization. 
For the prediction, we use a simple node-specific multi-layer perception (MLP).
This task can be framed as a simple $1$-layer graph neural network (GNN) architecture.
% following the message-passing scheme, where the interaction graph itself is a parameter and subject to the training objective. 

For an efficient solution to the \emph{graph inference problem}, we propose \texttt{GINA} (Graph Inference Network Architecture).
% \texttt{GINA}'s (i) simultaneous optimization of the interaction graph and the dynamics, (ii)  computationally simple neighborhood aggregation, (iii) efficient weight sharing mechanism, and (iv) differentiable graph representation render the \emph{graph inference problem} feasible for systems with hundreds of nodes. 
\texttt{GINA} tackles this problem by simultaneously optimizing the interaction graph and the dynamics.
The representation employs a computationally simple neighborhood aggregation and an efficient weight-sharing mechanism between nodes.
In combination with a differentiable graph representation, this enables the application of our methods to systems with hundreds of nodes.
% bisscchen zu hart die Aussage:
%We find that \texttt{GINA} identifies the ground truth graph with high accuracy for different types of dynamical process and graph types. Particularly, we beat statistical baselines (e.g., based on pair-wise mutual information) in the presence of complex processes.

%Naturally, there are some limitations. Noticeably, we consider arbitrary dynamical models but require that the dynamics manifest itself in a statistical association among neighboring nodes in a meaningful way. In other words, we define the interaction graph such that all direct interactions are represented as edges. We also assume a system, where all components behave reasonably similar to another. 
%The nature of the aggregation scheme makes the method also unsuitable for interact on graphs with different edge types. 
%Moreover, we identify sample size as a significant factor determining the reconstruction accuracy. 
%Lastly, we identify a trade-off between considering only pair-wise interactions (completely model-free), and joint interactions that take the whole neighborhood into account but constrains the types of interactions.

%In a broader sense, we examine how dynamical processes manifest themselves in temporal and spatial associations. For temporal association (i.e., when time series data is available), it is known that graph inference is widely possible. However, it remains underexplored in literate graph inference is even possible when only looking at independent snapshots, that is, if the information to recover the interaction topology is even theoretically present.  

In  summary, we formalize and test the hypothesis that network reconstruction can be formulated as a prediction task. This contribution includes: (i) we propose a suitable neighborhood aggregation mechanism, (ii) we propose the neural architecture \texttt{GINA} to efficiently solve the prediction and reconstruction task, and (iii) we evaluate \texttt{GINA} on synthetically generated snapshots using various combinations of graphs and diffusion models.

%%%%%%%%%%%%%%%%%%%%%%%%%%%%%%
%%%%%%%%%%%%%%%%%%%%%%%%%%%%%%
%  Related Work
%%%%%%%%%%%%%%%%%%%%%%%%%%%%%%
%%%%%%%%%%%%%%%%%%%%%%%%%%%%%%
\section{Related work} 
Literature abounds with methods to infer the (latent) functional organization of complex systems that is often expressed using (potentially weighted, directed, temporal, multi-) graphs.

Most relevant for this work are previous approaches that use deep learning on time series data to learn the interaction dynamics and the interaction structure. Zhang et al.~propose a two-component GNN architecture where a graph generator network propose interaction graphs and a dynamics learner learns to predict the dynamical evolution using the interaction graph \cite{zhang2019general,zhang2021automated}. Both components are trained alternately. 
Similarly, Kipf et al.~learn the dynamics using an encoder-decoder architecture that is constrained by an interaction graph which is optimized simultaneously \cite{kipf2018neural}. Huang et al.\ use a compression-based method for this purpose \cite{huang2020sdare}. 
Another state-of-the-art approach for this problem, based on regression analysis instead of deep learning, is the ARNI framework by Casadiego et al.~\cite{casadiego2017model}.
This method, however, requires time-series data and hinges on a good choice of basis functions.
% \paragraph{Casadiego et al.~\cite{casadiego2017model}}
% \begin{itemize}
%     \item time series data
% \end{itemize}
%In all cases differences are:  architecture, joint training.

Other methods to infer interaction structures aim at specific dynamical models and applications. Examples include epidemic contagion \cite{newman2018estimating,di2020network,prasse2020network}, gene networks \cite{kishan2019gne,omranian2016gene}, functional brain network organization \cite{de2018connectivity}, and protein-protein interactions \cite{hashemifar2018predicting}. 
In contrast, our approach assumes no prior knowledge about the laws that govern the system's evolution.

Statistical methods provide an equally viable and often very robust alternative. 
These can be based on partial correlation, mutual information, or graphical lasso \cite{tibshirani1996regression,friedman2008sparse}.
Here, we not only rely on pair-wise correlations among components but also the joint impact of all neighboring components, which 
is necessary in the presence of non-linear dynamical laws governing the system.
 Moreover, we directly infer binary (i.e., unweighted) graphs in order to not rely on (often unprincipled) threshold mechanisms.

%\subsection{Graph neural networks} 
%\subsubsection{Edge Prediction}
%\subsubsection{Graph Generation}

%%%%%%%%%%%%%%%%%%%%%%%%%%%%%%
%%%%%%%%%%%%%%%%%%%%%%%%%%%%%%
%  Problem Setting
%%%%%%%%%%%%%%%%%%%%%%%%%%%%%%
%%%%%%%%%%%%%%%%%%%%%%%%%%%%%%
\section{Foundations and problem formulation}
\begin{figure}[t]
\centering
\includegraphics[width=0.98\textwidth ]{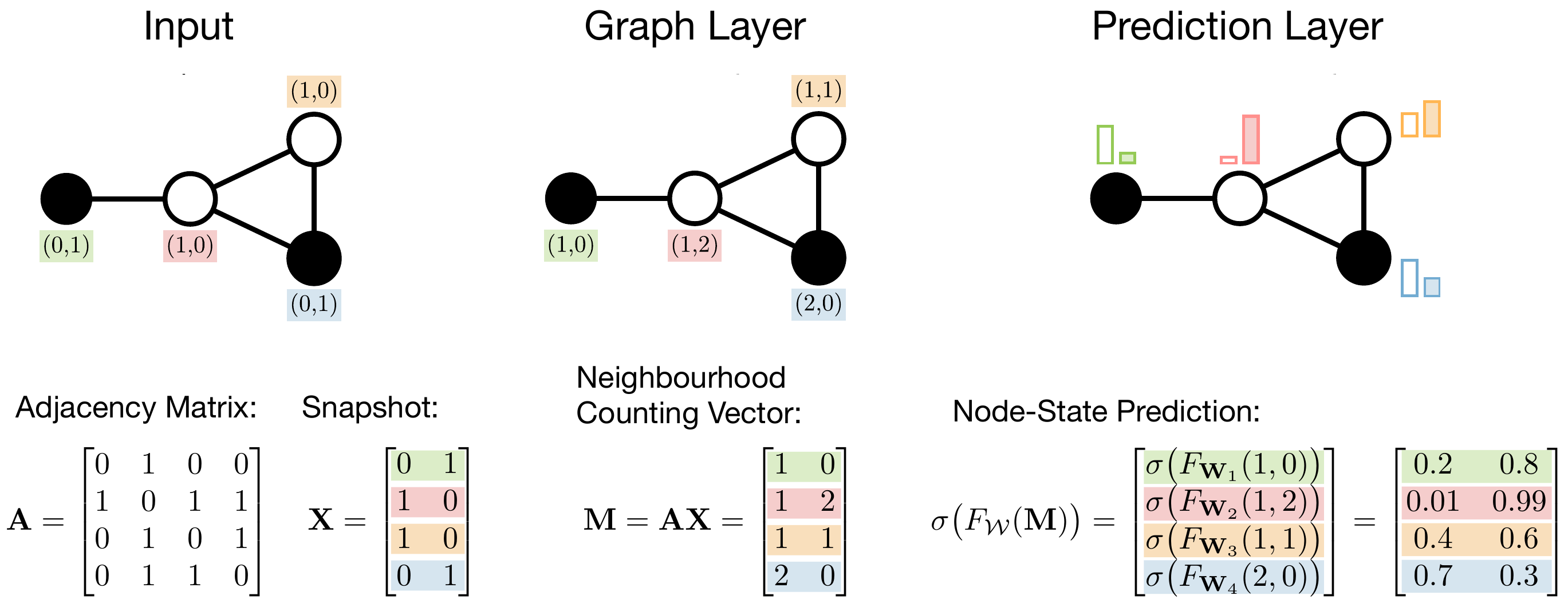}
\caption{Schematic architecture using $4$-node graph with $\mathcal{S} = \{(0,1),(1,0)\}$. Nodes are color-coded, node-states are indicated by the shape (filled: I, blank: S). First, we compute $m_i$ for each node $v_i$ (stored as $\textbf{M}$), then we feed each $m_i$ into a predictor that predicts the original state. }
 \label{fig:notation}
\end{figure}

The goal is to find the latent interaction graph of a complex system with $n$ agents/nodes. A graph is represented as an adjacency matrix $\mathbf{A}$ of dimension $n \times n$  (with node set $\{v_i \mid 1 \leq i \leq n \}$), where $a_{ij} \in \{0,1\}$ indicates the presence ($a_{ij}=1$) or absence ($a_{ij}=0$) of an edge between node $v_i$ and $v_j$. We assume that $\mathbf{A}$ is symmetric (the graph is undirected) and the diagonal entries are all zero (the graph has no self-loops).  
We use $\mathbf{A}^*$ to denote the ground truth matrix. 
%Note that we assume an ordering of nodes. This makes it possible to distinguish isomorphic graphs. The node ordering is implicitly given by the snapshots.

Each snapshot assigns a node-state to each node.
The finite set of possible node-states is denoted $\mathcal{S}$.
For convenience, we assume that the node-states are represented using one-hot encoding.
For instance, in an epidemic model nodes might be susceptible or infected.
Since there are two node-states, we use $\mathcal{S} = \{(0,1),(1,0)\}$. 
Each snapshot $\mathbf{X} \in  \{0,1\}^{n \times |\mathcal{S}| }$ can then conveniently be represented as a matrix with $n$ rows, where each row describes the corresponding one-hot encoded node-state (cf.~Fig.~\ref{fig:notation} left). 
We use $\mathcal{X}$ to denote the set of independent snapshots.
We make no specific assumption about the underlying distribution or process behind the snapshots or their relationship to another.
%Moreover, we can process each snapshot individually.  

For a node $v_i$ (and fixed snapshot), we use $m_i \in \mathbb{Z}^{|\mathcal{S}|}_{\geq 0} $ to denote the (element-wise) sum of all neighboring node-states, referred to as \emph{neighborhood counting vector}.
For example, consider again the case $\mathcal{S} = \{(1,0),(0,1)\}$:
$m_i=(10,13)$ refers to a node $v_i$ with $10$ susceptible ($(1,0)$) and $13$ infected ($(0,1)$) neighbors.
Note that the sum over $m_i$ is the degree (i.e., number of neighbors) of that node (here, $10+13=23$).
The set of all possible neighborhood counting vectors is denoted by $\mathcal{M} \subset \mathbb{Z}^{|\mathcal{S}|}_{\geq 0}$.
We can compute neighborhood counting vectors of all nodes using $\mathbf{M} = \mathbf{A}\mathbf{X}$, where the $i$-th row of $\mathbf{M}$ equals $m_i$ (cf.~Fig.~\ref{fig:notation} center).

In the next step, we feed each counting vector $m_i$ in a machine learning model that predicts the original state of $v_i$ in that snapshot.
Specifically, we learn a function $F_{\mathbf{W}_i}: \mathcal{M} \rightarrow \texttt{Distr}(\mathcal{S})$, where $\texttt{Distr}(\mathcal{S})$ denotes the set of all probability distributions over $\mathcal{S}$ (in the sequel, we make the mapping to a probability distribution explicit by adding a $\texttt{Softmax}$ function). 
%$F_i$ learns to predict the state of a node, given the (aggregated) node-states of its neighbors. 
%To evaluate $F_i$ and $\mathbf{A}$, we can simply predict the state of each node in each snapshot. 
To evaluate $F_{\mathbf{W}_i}$, we use some loss function to quantify how well the distribution predicts the true state and minimize this \emph{prediction loss}.
%The goal is to optimize $\mathbf{A}$ and $F_i$ (for all $i$) such that this  is minimized. 
We assume that $F_{\mathbf{W}_i}$ is fully parameterized by a node-dependent weight matrix $\mathbf{W}_i$. 
%Hence, we use $F_{\mathbf{W}_i}$ instead of $F_i$ in the sequel.
All weights in a network are given by the list $\mathcal{W}=\{\mathbf{W}_1, \dots, \mathbf{W}_n\}$. 
We also define 
${F}_{\mathcal{W}}(\cdot)$ as a node-wise application of $F_{\mathbf{W}_i}(\cdot)$, that is, 
\begin{align*}
F_{\mathcal{W}}(m_1, m_2, \dots, m_n) = (F_{\mathbf{W}_1}(m_1), F_{\mathbf{W}_2}(m_2), \dots ,F_{\mathbf{W}_n}(m_n)) \;\;.
\end{align*}

The hypothesis is that the ground truth adjacency matrix $\mathbf{A}^*$ provides the best foundation for $F_{\mathbf{W}_i}$, ultimately leading to the smallest prediction loss.
Under this hypothesis, we can use the loss as a surrogate for the accuracy of candidate $\mathbf{A}$ (compared to the ground truth graph). 
The number of possible adjacency matrices of a system with $n$ nodes (assuming no self-loops and symmetries) is  $2^{n(n-1)/2}$. 
%(the exponent is the size of an upper triangular $n \times n$ matrix without the diagonal). 
Thus, it is hopeless to simply try \emph{all} possible adjacency matrices.  
Hence, we need to strengthen this hypothesis and assume that smaller distances between $\mathbf{A}$ and $\mathbf{A}^*$ (we call this \emph{graph loss}) lead to smaller prediction losses (in a reasonable sense).
This way, the optimization becomes feasible and we can follow the surrogate loss in order to arrive at $\mathbf{A}^*$. 

\paragraph{Graph neural network}
Next, we formulate the \emph{graph inference problem} using a graph neural network $N(\cdot)$ that loosely resembles an autoencoder architecture: In each snapshot $\textbf{X}$, we predict the node-state of each node using only the neighborhood of that node. Then, we compare the prediction with the actual (known) node-state.

For a given adjacency matrix (graph) $\mathbf{A} \in \{0,1\}^{n\times n} $
and list of weight matrices $\mathcal{W}$, we define the GNN $N(\cdot)$, applied to a snapshot $\mathbf{X}  \in \{0,1\}^{n \times |\mathcal{S}|} $ as:
\begin{align*} 
&{N}_{\mathcal{W},\mathbf{A}}:  \{0,1\}^{ n\times |\mathcal{S}|} \rightarrow \mathbb{R}^{n \times |\mathcal{S}|} \\
&{N}_{\mathcal{W},\mathbf{A}}: \mathbf{X}  \mapsto   \texttt{Softmax} \big(F_{\mathcal{W}}( \mathbf{A}\mathbf{X})\big) \;
\end{align*}
where $\texttt{Softmax} (\cdot)$ is applied row-wise.
Thus, ${N}_{\mathcal{W},\mathbf{A}}(\mathbf{X})$ results in a matrix where each row corresponds to a node and models a distribution over node-states.
Similar to the auto-encoder paradigm, input and output are of the same form and the network learns to minimize the difference (note that the one-hot encoding can also be viewed as a valid probability distribution over node-states). The absence of self-loops in $\mathbf{A}$ is critical as it means that the actual node-state of a node is not part of their own neighborhood aggregation.
As we want to predict a node's state, the state itself cannot be part of the input.  
%Otherwise, nodes would predict their own states, which would be meaningless.

We will refer to the matrix multiplication $\mathbf{A}\mathbf{X}$ as \emph{graph layer} and to the application of $ \texttt{Softmax}(F_{\mathcal{W}}( \cdot))$ as \emph{prediction layer}.
We only perform a single application of the graph layer on purpose, which means that only information from the immediate neighborhood can be used to predict a node-state.
% Using $n$-hop neighborhoods would make the network more powerful but defeat the purpose of graph reconstructing. 
While using $n$-hop neighborhoods would increase the network's predictive power it would be detrimental to graph reconstruction.

Most modern GNN architectures are written as in the message-passing scheme, where each layer performs an \emph{aggregate}$(\cdot)$ and a \emph{combine}$(\cdot)$ step. 
The \emph{aggregate}$(\cdot)$ step computes a neighborhood embedding based on a permutation-invariant function of all neighboring nodes. The \emph{combine}$(\cdot)$ step combines this embedding with the actual node-state. 
In our architecture, aggregation is the element-wise sum. The combination, however, needs to purposely ignore the node-state (in order for the prediction task to make sense) and applies $ \texttt{Softmax}(F_{\mathcal{W}}( \cdot))$. 

\paragraph{Prediction loss}
We assume a loss function $L$ that is applied independently for each snapshot:
\begin{align*}
&L:    \{0,1\}^{ n\times |\mathcal{S}|} \times \mathbb{R}^{n \times |\mathcal{S}|} \rightarrow \mathbb{R} 
%&L: \mathbf{X}, {N}_{\mathcal{W},\mathbf{A}} \mapsto L \big( \mathbf{X}, {N}_{\mathcal{W},\mathbf{A}} \big) 
\end{align*}

The prediction loss $L$ compares the input (actual node-states) and output (predicted node-states) of the GNN  $N$. 
We define the loss on a set of independent snapshots $\mathcal{X}$ as the sum over all constituent snapshots:
\begin{align*}
    L \big(\mathcal{X}, {N}_{\mathcal{W},\mathbf{A}}(\mathcal{X}) \big) \coloneqq  \sum_{\mathbf{X} \in \mathcal{X}} L \big(\mathbf{X},  {N}_{\mathcal{W},\mathbf{A}} (\mathbf{X}) \big) \;.
\end{align*}
In our experiments, we use row-wise MSE-loss. 

Note that, in the above sum, all snapshots are treated equally independent of their corresponding initial conditions or time points at which they were made (which we do not know anyway). 
%This reflects the assumption that all arising patterns  are the result of local interactions which are constrained by the  underlying graph structure.

\paragraph{Graph inference problem}

We define the  \emph{graph inference problem} as follows:
For given set of snapshots $\mathcal{X} = \{ \mathbf{X}_1, \dots , \mathbf{X}_m \}$ (corresponding to $n$ nodes), find an adjacency matrix $\mathbf{A}' \in \{0,1\}^{n\times n} $ and list of weight matrices $\mathcal{W}'$ minimizing the prediction loss:
\begin{align*}
(\mathcal{W}', \mathbf{A}') \coloneqq \argmin_{\mathcal{W}, \mathbf{A}} L \big(\mathcal{X}, {N}_{\mathcal{W},\mathbf{A}}(\mathcal{X}) \big) \,.
\end{align*}
% is minimized. 
Note that, in general we cannot guarantee that $\mathbf{A}'$ is equal to the ground truth matrix $\mathbf{A}^*$. 
Regarding the computational complexity, it is known that network reconstruction for epidemic models based on time series data is $\mathcal{NP}$-hard when formulated as a decision problem \cite{prasse2018maximum}. We believe that this carries over to our setting but leave a proof for future work. 

%We also want to note that it is easy to construct pathological counter-examples where the solution $\mathbf{A}$ to the graph inference problem is very different from the actual ground truth graph $\mathbf{A}*$ or where there exists 

%%%%%%%%%%%%%%%%%%%%%%%%%%%%%%
%%%%%%%%%%%%%%%%%%%%%%%%%%%%%%
% GINA
%%%%%%%%%%%%%%%%%%%%%%%%%%%%%%
%%%%%%%%%%%%%%%%%%%%%%%%%%%%%%
\section{Our method: \texttt{GINA} \label{sec:Gina}}

\begin{figure}[t]
\centering
\includegraphics[width=0.55\textwidth ]{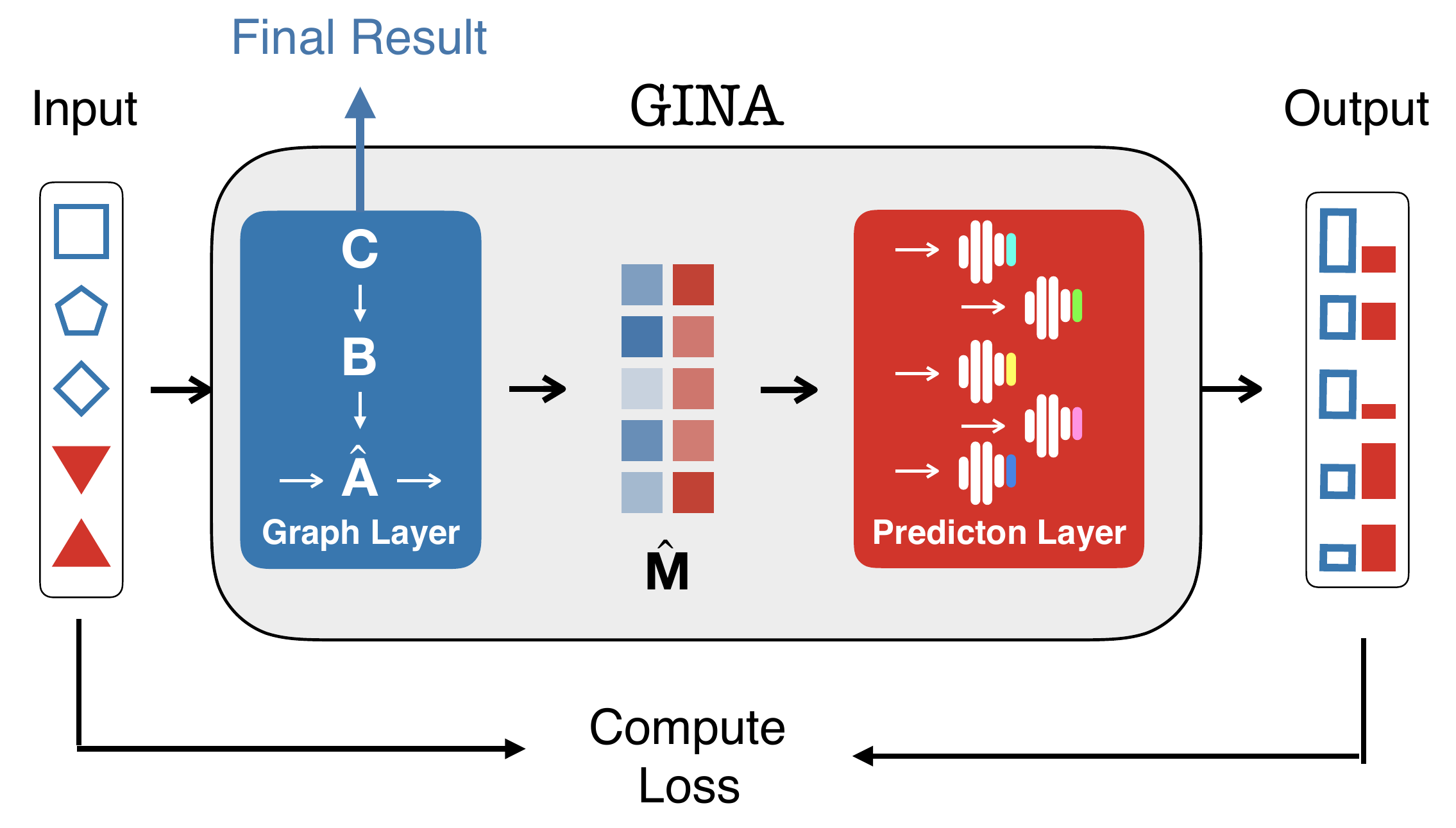}
\caption{Illustration of \texttt{GINA}. Snapshots are processed independently. Each input/snapshot associates each node with a state and each output with a distribution over states. The training process optimizes the distribution. 
The output is computed based on a multiplication with the current adjacency matrix candidate (stored as $\textbf{C}$) and the application of a node-wise MLP. Ultimately, we are interested in a binarized version of the adjacency matrix. Color/filling indicates the state, shape identifies nodes.}
 \label{fig:gina}
\end{figure}

As explained in the previous section, it is not possible to solve the graph inference problem by iterating over all possible graphs/weights. 
Hence, we propose \texttt{GINA} (Graph Inference Network Architecture). \texttt{GINA}  efficiently approximates the graph inference problem by jointly optimizes the graph $\mathbf{A}$ and the predictor layer weights $\mathcal{W}$.

Therefore, we adopt two tricks: Firstly, we impose a relaxation on the graph adjacency matrix representing its entries as real-valued numbers. Secondly, we use shared weights in the weight matrices belonging to different nodes.
%$\mathcal{W}=\{\mathbf{W}_1, \dots, \mathbf{W}_n\}$.
Specifically, each node $v_i$ gets its custom MLP, but weights of all layers, except the last one, are shared among nodes.
%Note that there is a natural trade-off between efficiency and allowed differences among node-dependent dynamics. 
This allows us to simultaneously optimize the graph and the weights using back-propagation. 
Apart from that, we still follow the architecture from the previous section. That is, a network layer maps a snapshot to neighborhood counting vectors, and each neighborhood counting vector is pushed through a node-based MLP.

\paragraph{Graph layer}
Internally, we store the interaction graph as an upper triangular matrix $\textbf{C}$.
In each step, we (i) compute $\textbf{B} = \textbf{C} + \textbf{C}^\top$ to enforce symmetry, (ii) compute $\hat{\mathbf{A}} = \mu(\textbf{B})$, and (iii) set all diagonal entries of $\hat{\mathbf{A}}$ to zero.
Here, $\mu$ is a differential function that is applied element-wise and maps real-valued entries to the interval $[0,1]$. It ensures that $\hat{\mathbf{A}}$ approximately behaves like a binary adjacency matrix while remaining differentiable (the hat-notation indicates relaxation). 
Specifically, we adopt a Sigmoid-type function $f$ that is parametrized by a sharpness parameter $v$:
\begin{align*}
&\mu: \mathbb{R} \rightarrow [0,1] \\
& \mu: x \mapsto  f \Big(  \big(f(x)-0.5 \big) \cdot v \Big)    \;,
\end{align*}
where we choose $f(x)=1/(1+ \texttt{exp}(-x))$ and increase the sharpness of the step using $v$ over the course of the training.
Finally, the graph layer matrix is multiplied with the snapshot (i.e., $\hat{\mathbf{M}} = \hat{\mathbf{A}}\textbf{X}$) and yields a relaxed version of the neighborhood counting vectors.

\paragraph{Prediction layer}
In  $\hat{\mathbf{M}}$, each row corresponds to one node.  Thus, we apply the MLPs independently to each row. We use $\hat{m}_i$ to denote the row corresponding to node $v_i$ (i.e., the neighborhood counting relaxation of $v_i$). 
Let $FC_{i,o}$ denote a fully-connected (i.e., linear) layer with input (resp.\;output) dimension $i$ (resp.\;$o$). We use a $\texttt{ReLu}$ and Softmax activation function. The whole prediction MLP contains four sub-layers and is given as:
\begin{align*}
&o^1_i  = \texttt{ReLu} (FC_{|\mathcal{S}|,10}(\hat{m}_i)) \\
&o^2_i  = \texttt{ReLu} (FC_{10,10}(o^1)) \\
&o^3_i  = \texttt{Softmax} (FC_{10,|\mathcal{S}|}(o^2)) \\
&o^4_i  =  \texttt{Softmax} (FC_{|\mathcal{S}|,|\mathcal{S}|}(o^3)) \;.
\end{align*}

Only the last sub-layer contains node-specific weights. This layer enables a node-specific shift of the probability computed in the previous layer. Note that we use a comparably small dimension (i.e., 10) for internal embeddings, which has shown to be sufficient in our experiments.

\paragraph{Training}
We empirically find that over-fitting is not a problem and therefore do not use a test set.
A natural approach would be to split the snapshots into a training and test set and optimize $\hat{\mathbf{A}}$ and $\mathcal{W}$ on the training set until the loss reaches a minimum on the test set.
Specifically, the loss on the test set provides the best surrogate for the distance to the ground truth graph.
Another important aspect during training is the usage of mini-batches. For ease of notation, we ignored batches so far.
In practice, mini-batches are crucial for fast and robust training.
A mini-batch of size $b$, can be created by concatenating $b$ snapshots (in the graph layer) and re-ordering the rows accordingly (in the prediction layer).
%For more information, we refer the reader to the documentation of the prototype implementation.    

\paragraph{Limitations}
There are some relevant limitations to \texttt{GINA}. Firstly, we can provide no guarantees that the ground truth graph is actually the solution to the \emph{graph inference problem}.
In particular, simple patterns in the time domain (that enable trivial graph inference using time series data) might correspond to highly non-linear patterns inside a single snapshot.
Moreover, \texttt{GINA} is only applicable if statistical associations among adjacent nodes manifest themselves in a way that renders the counting abstraction meaningful.
Statistical methods are more robust in the way they can handle different types of pair-wise interactions but less powerful regarding non-linear combined effects of the complete neighborhood.
Another relevant design decision is to use one-hot encoding which renders the forward pass extremely fast but will reach limitations when the node-state domain becomes very complex.
%This is new:
Together with relational homogeneity, we also assume that all agents behave reasonably similar to another which enables weight sharing and therefore greatly increases the efficiency of the training and reduces the number of required samples. 

\section{Experiments\label{sec:experiments}}

%
% examples Snapshots
%
\begin{figure}[t]
\centering
\includegraphics[height=2.8cm]{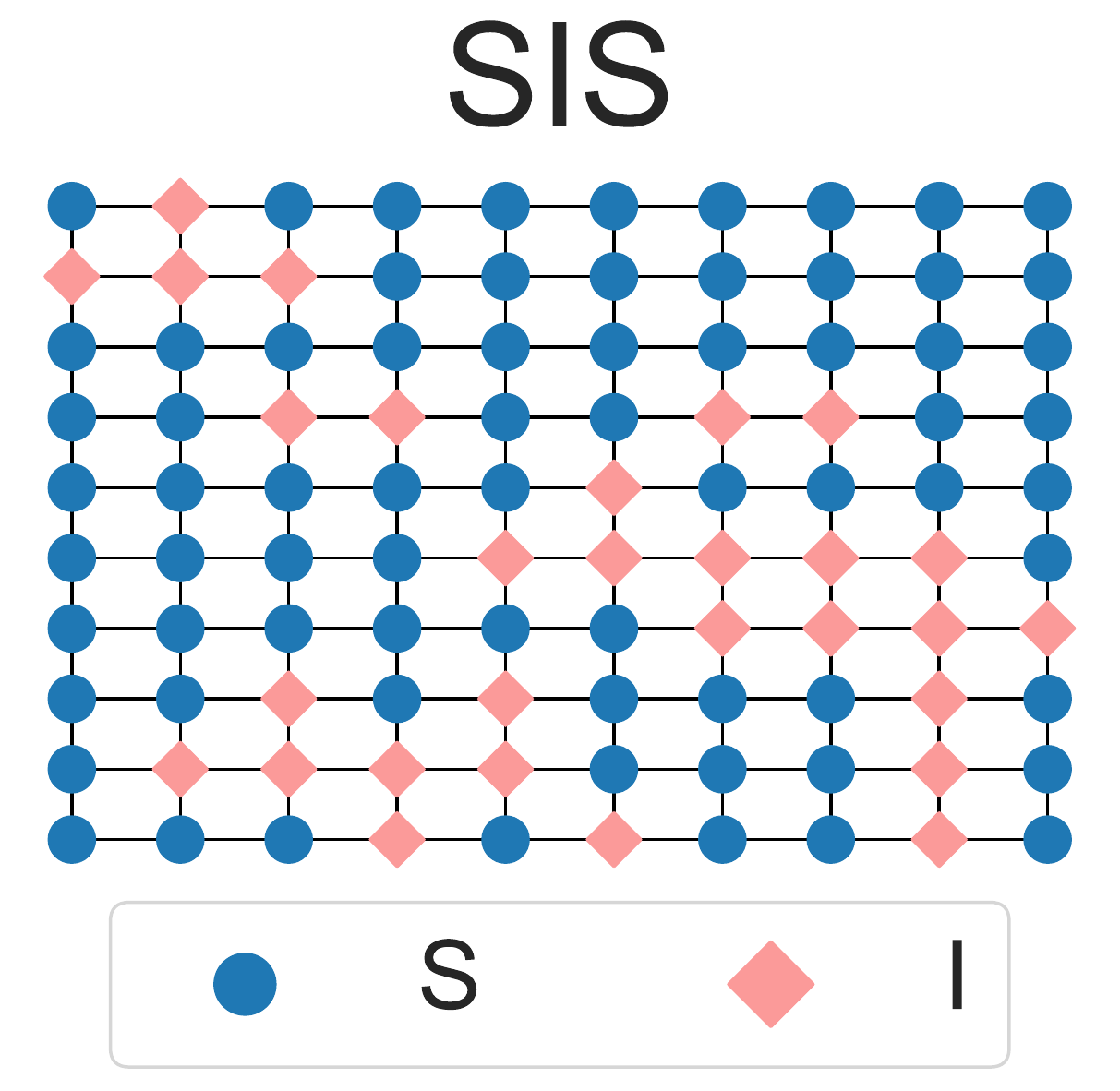}
\phantom{x}
\includegraphics[height=2.8cm]{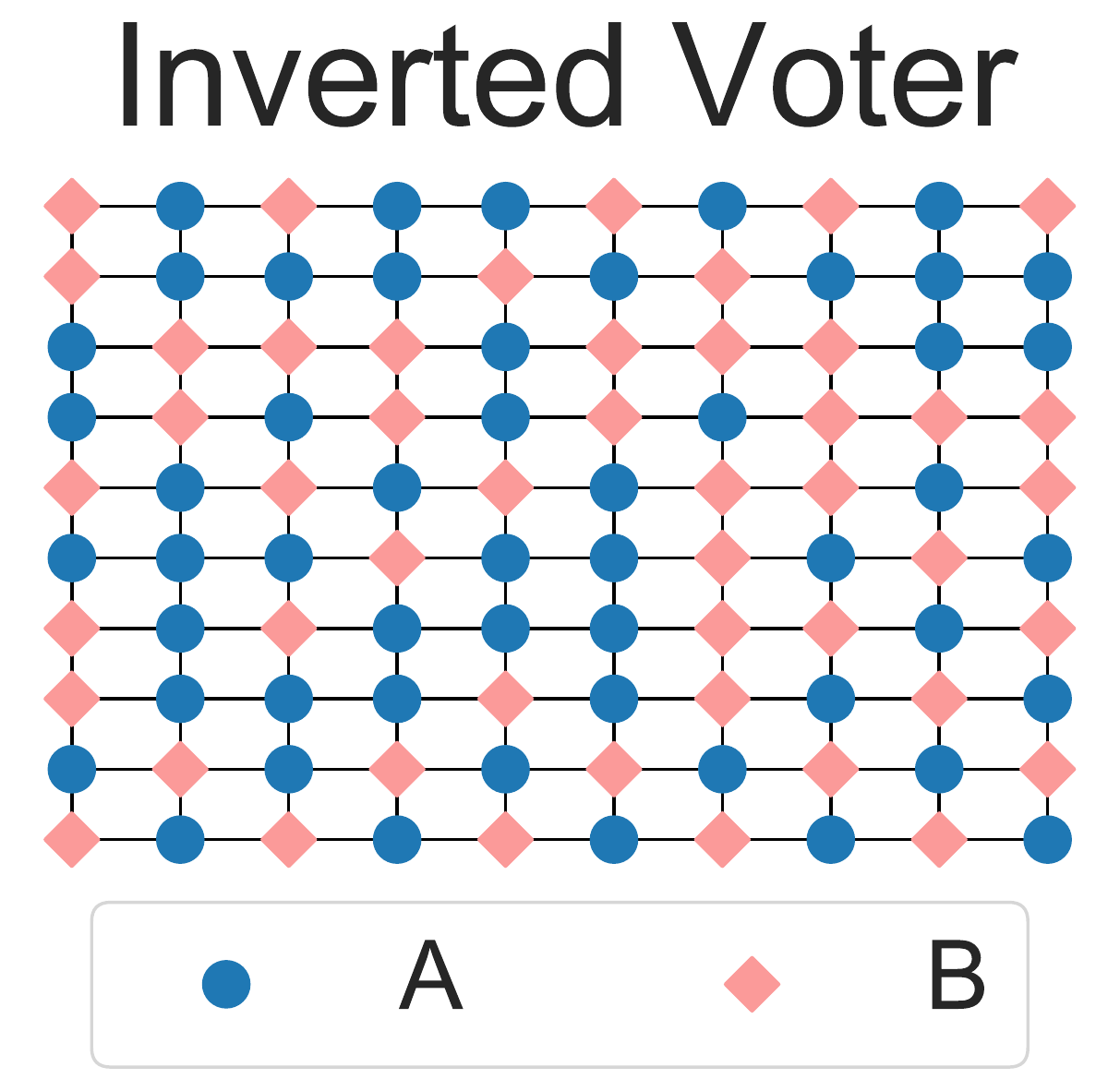}
\phantom{x}
\includegraphics[height=2.8cm]{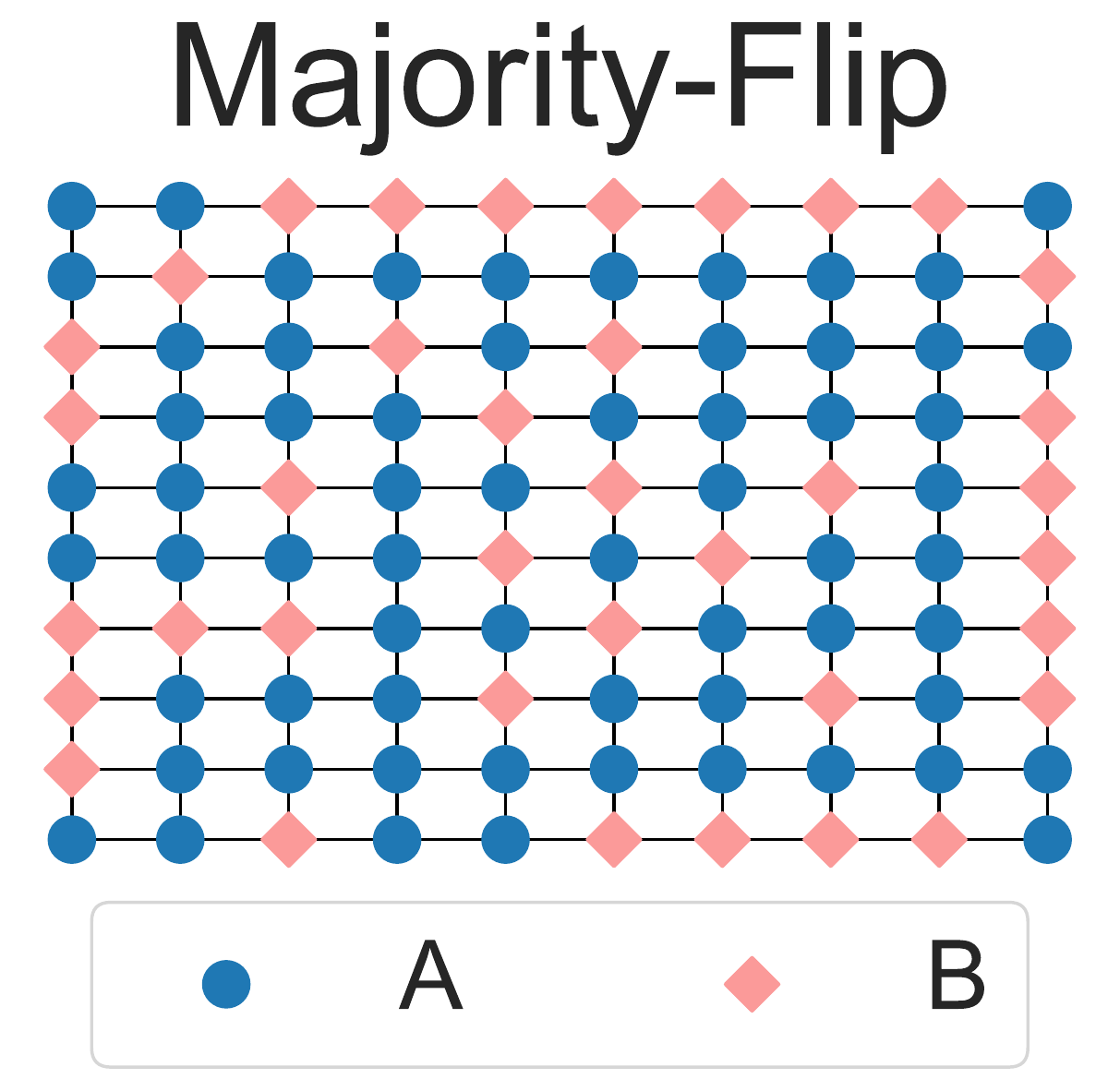}
\includegraphics[height=2.8cm]{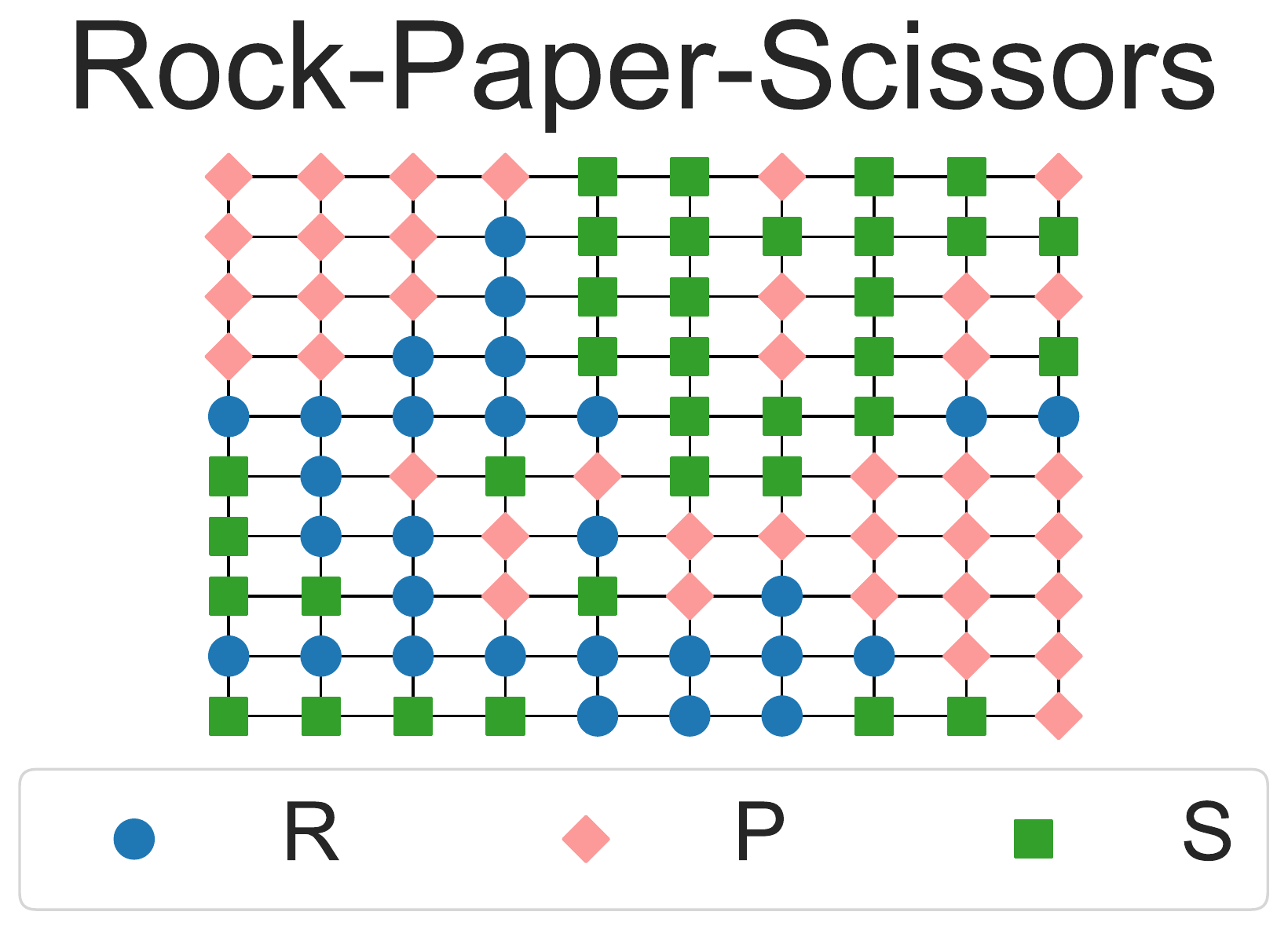}
\caption{\textbf{Exp.~2:} Examples of typical equilibrium snapshots on a $10 \times 10$ grid graph. Different dynamics give rise to different types of cluster formations. }
 \label{fig:exp3snapshots}
\end{figure}

We conduct three experiments using synthetically generated snapshots. In \textbf{Experiment 1}, we analyze the underlying hypothesis that the ground truth graph enables the best node-state prediction. In \textbf{Experiment 2}, we study the importance of sample size for the reconstruction accuracy, and in \textbf{Experiment 3}, we compare \texttt{GINA} to statistical baselines. 
Our prototype of \texttt{GINA} is implemented using PyTorch \cite{paszke2019pytorch} and is executed on a standard desktop computer with an {NVIDIA GeForce RTX 3080}, 32 GB of RAM, and an {Intel i9-10850K} CPU. Open source code (GNU GPLv3) is available at GitHub\footnote{\hyperlink{https://github.com/GerritGr/GINA}{github.com/GerritGr/GINA}}.

To measure the quality of an inferred graph, we compute the distance to the ground truth graph. We define this \emph{graph loss} as the $L_1$ (Manhattan) distance of the upper triangular parts of the two adjacency matrices (i.e., the number of edges to add/remove). We always use a binarized version of $\hat{\textbf{C}}$ for comparison with $\textbf{A}^*$.
All results are based on a single run of \texttt{GINA}, performing multiple runs and using the result with the lower prediction loss, might further improve \texttt{GINA}'s performance.  

\paragraph{Dynamical models}
We study six models. A precise description of dynamics and parameters are provided in Appendix \ref{sec:dynmodels}.
We focus on stochastic processes, as probabilistic decisions and interactions are essential for modeling uncertainty in real-world systems.
The models include a simple \textbf{SIS}-epidemic model where infected nodes can randomly infect susceptible neighbors or become susceptible again.
In this model, susceptible nodes tend to be close to other susceptible nodes and vice versa. This renders network reconstruction comparably simple.
In contrast, we also propose an Inverted Voter model (\textbf{InvVoter}) where nodes tend to maximize their disagreement with their neighborhood (influenced by the original Voter model\cite{campbell1954voter}) (cf.~Fig.~\ref{fig:exp3snapshots}). Nodes have one of two opinions (A or B) and nodes in A tend to move to B faster the higher their number of A neighbors and vice versa. To study the emergence of an even more complex pattern, we propose the \textbf{Majority-Flip} dynamics where nodes tend to change their current state when the majority of their neighbors follows the same opinion (regardless of the node's current state).  We refer the reader to Fig.~\ref{fig:predictOut} for a visualization of the node-state prediction conditioned.
%Living conditions are good (i.e., nodes tend to become/stay alive) when roughly half of an agent's neighbors are alive. 
We also examine a rock-paper-scissors (\textbf{RPS}) model to study evolutionary dynamics \cite{szabo2007evolutionary} and the well-known \textbf{Forest Fire} model \cite{bak1990forest}.% where a spot/node can be empty, occupied by a tree, or by fire induced by stochastic lightning. 
Finally, we test a deterministic discrete-time dynamical model: a coupled map lattice model (\textbf{CML}) \cite{garcia2002coupled,kaneko1992overview,zhang2019general} to study graph inference in the presence of chaotic behavior.
As the CML model admits real node-values  $[0,1]$, we performed discretization into $10$ equally-spaced bins.
For the stochastic models, we use numerical simulations to sample from the equilibrium distribution of the systems. For CML, we randomly sample an initial state and simulate it for a random time period. We do not explicitly add measurement errors but all nodes are subject to internal noise.

\paragraph{Experiment 1: Loss landscape}
\begin{figure}[t]
\centering
\includegraphics[height=2.2cm]{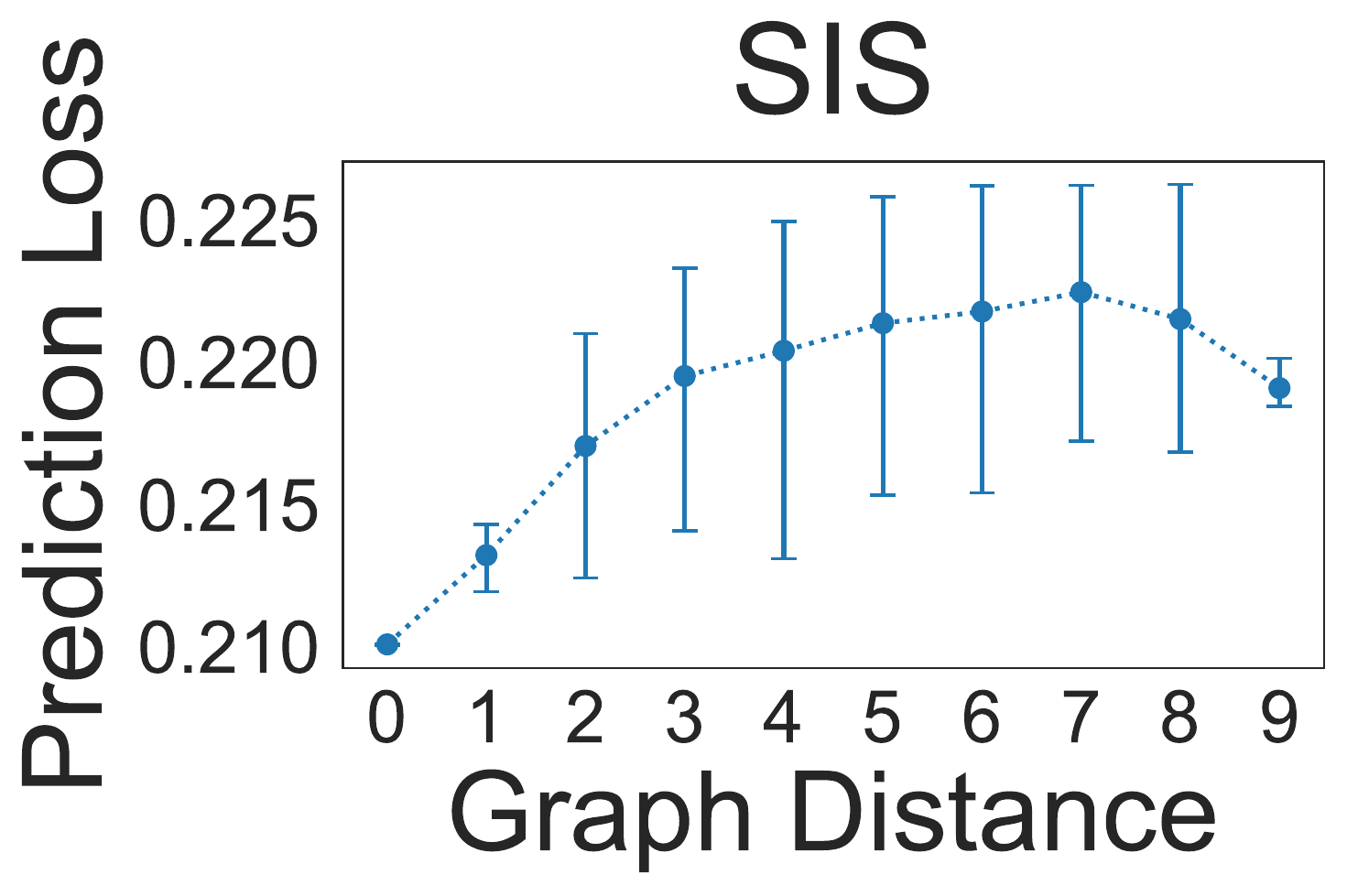}
\includegraphics[height=2.2cm]{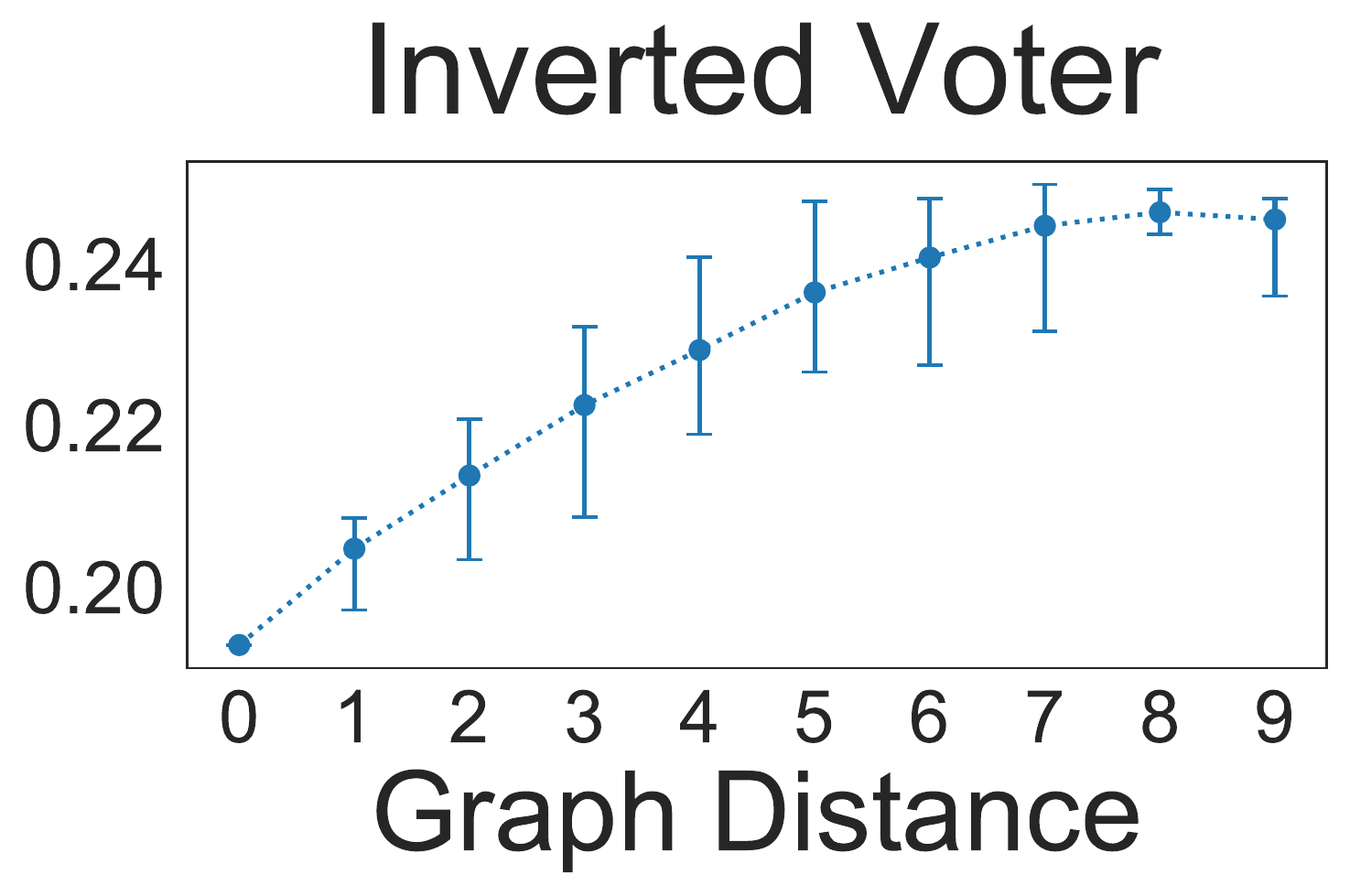}
\includegraphics[height=2.2cm]{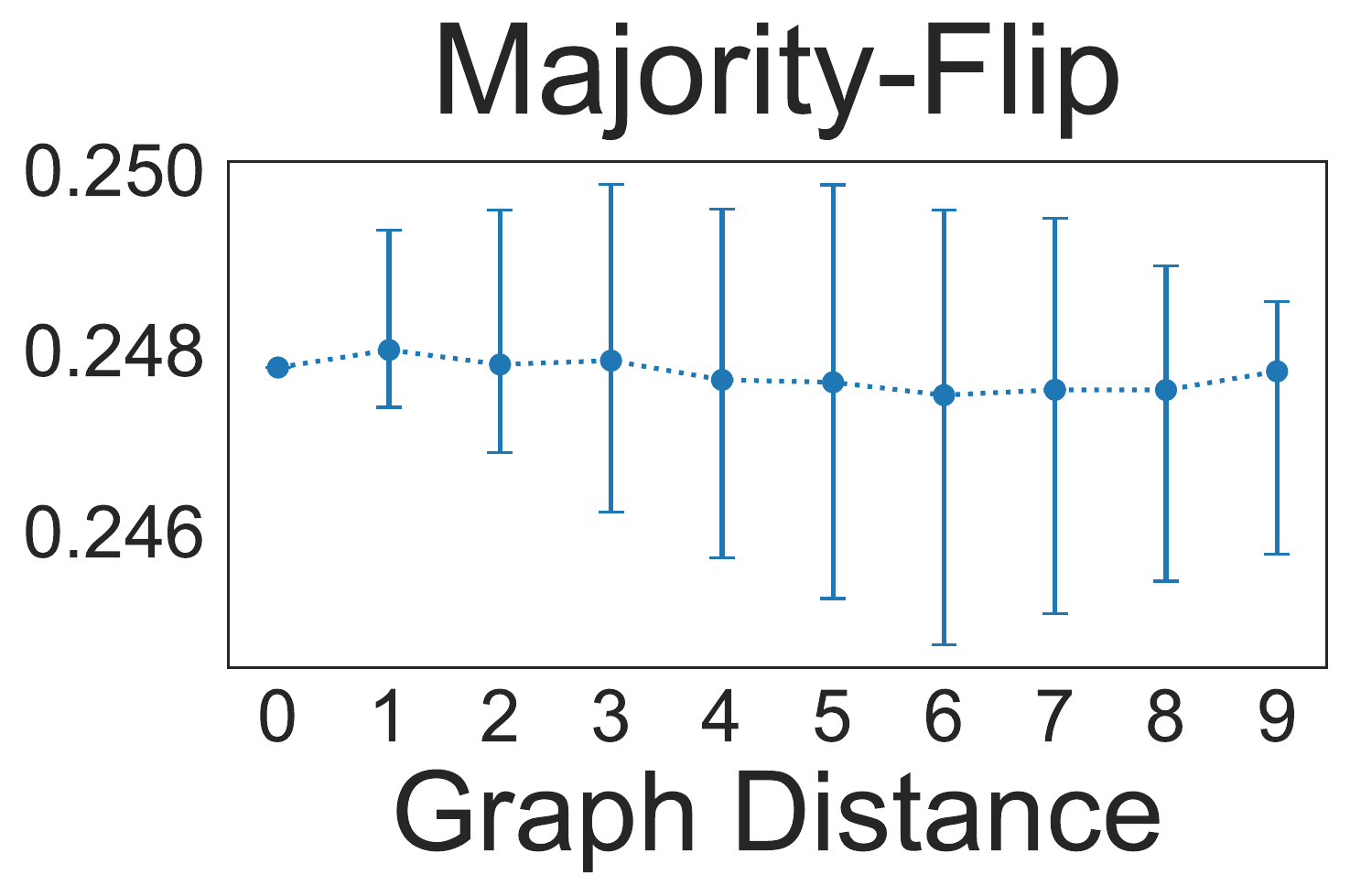}
\includegraphics[height=2.2cm]{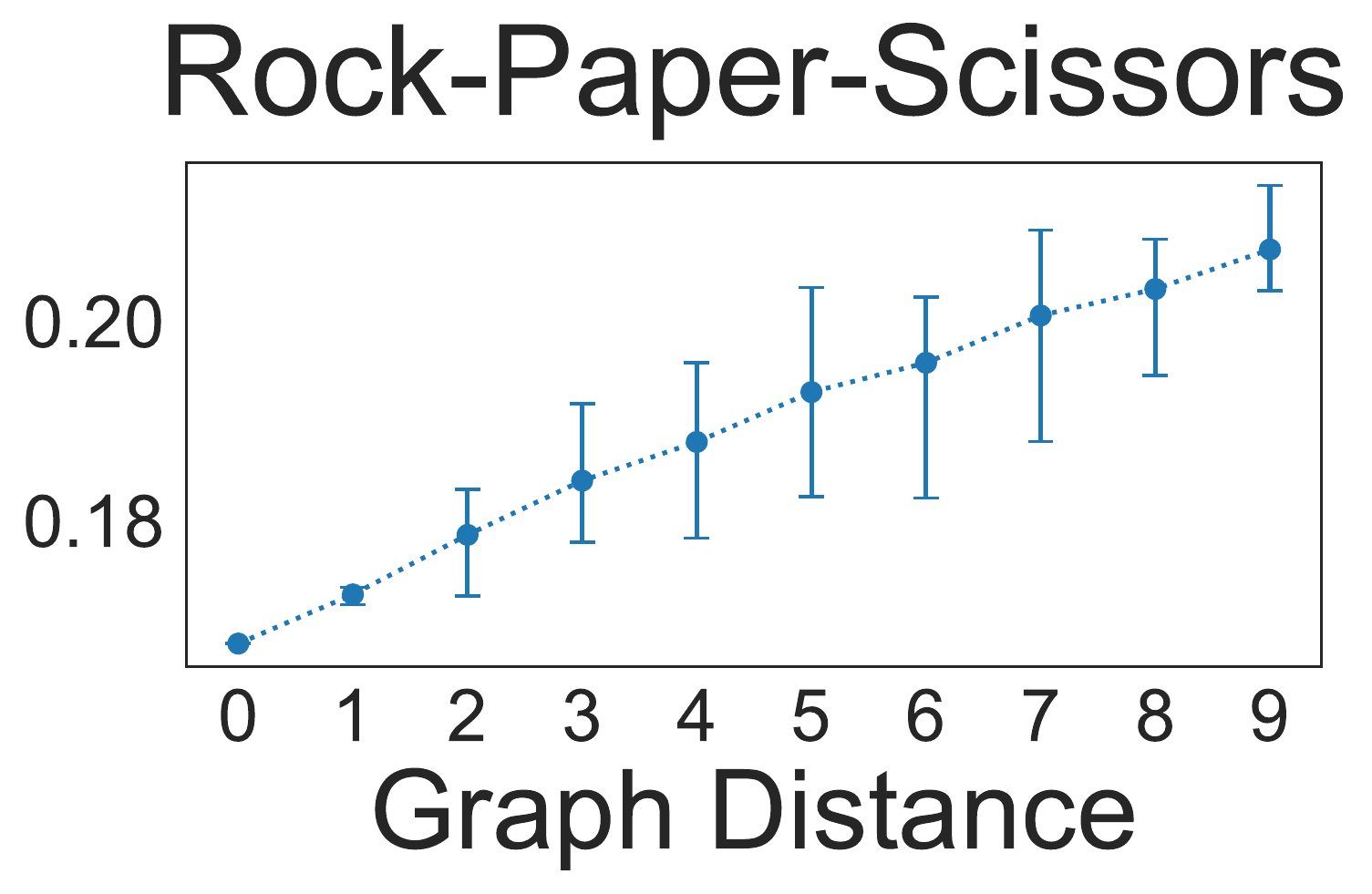}
\caption{\textbf{Exp.~1:} Computing the loss landscape based on all possible $5$-node graphs and $20000$ snapshots. $x$-axis: Graph distance to ground truth adjacency matrix. $y$-axis: Mean prediction loss of corresponding graph candidates. Error bars denote min/max-loss. }
 \label{fig:exp1}
\end{figure}

For this experiment, we generated snapshots corresponding to $4$ dynamical models on a the so-called \emph{bull graph} (as illustrated in Fig.\ \ref{fig:overview}). We then trained the network and measured the prediction loss for all potential all $5 \times 5$ adjacency matrices that represent connected graphs. We observe a large dependency between the prediction loss of a candidate matrix and the corresponding graph loss except for the Majority-Flip model.
Surprisingly, on larger graphs \texttt{GINA} still finds the ground truth graph with high accuracy given Majority-Flip snapshots compared to the baselines.

%distance to the ground truth matrix.

\paragraph{Experiment 2: Sample size}
\begin{figure}[t]
\centering
\phantom{xx}
\includegraphics[height=2.2cm]{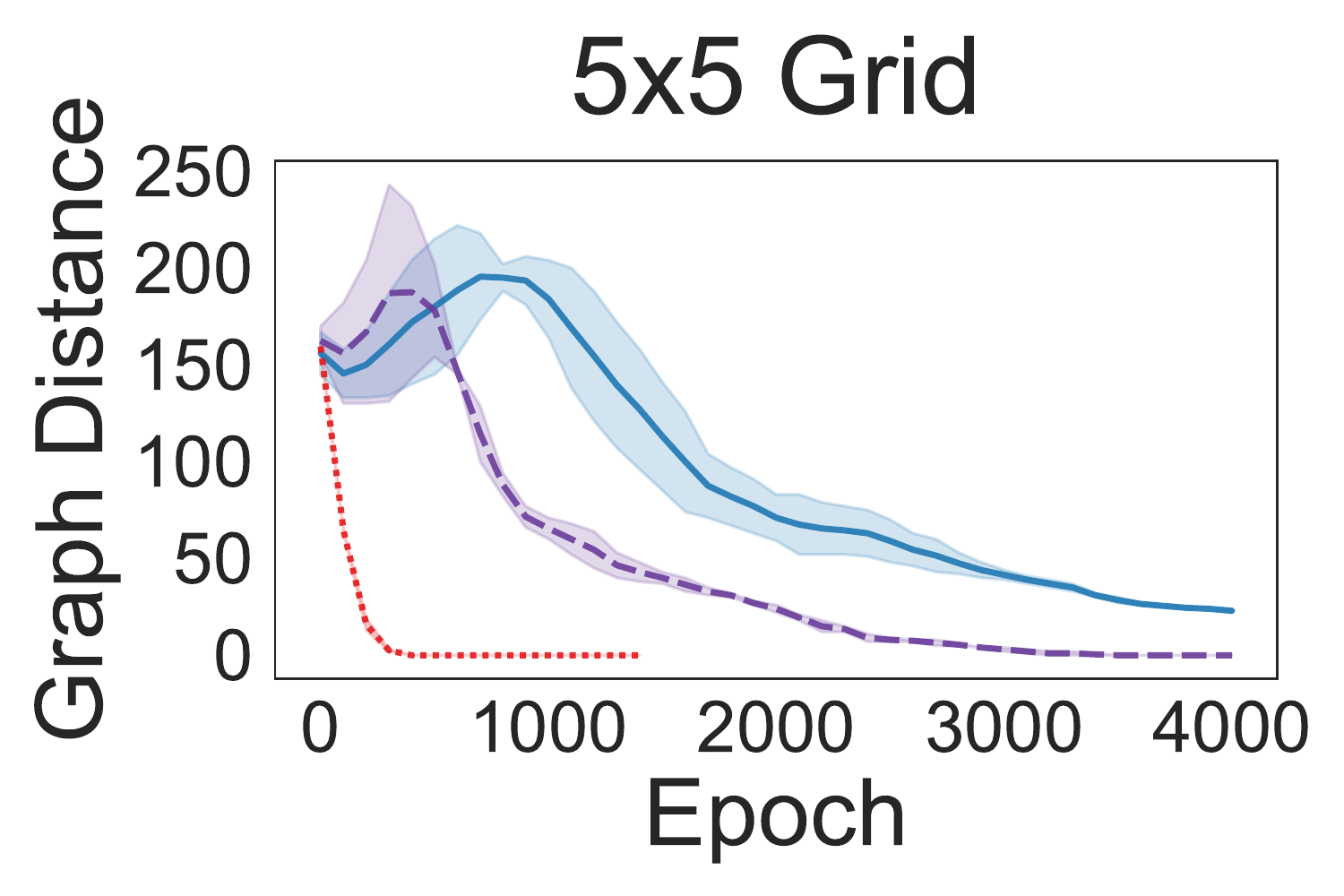}
\includegraphics[height=2.2cm]{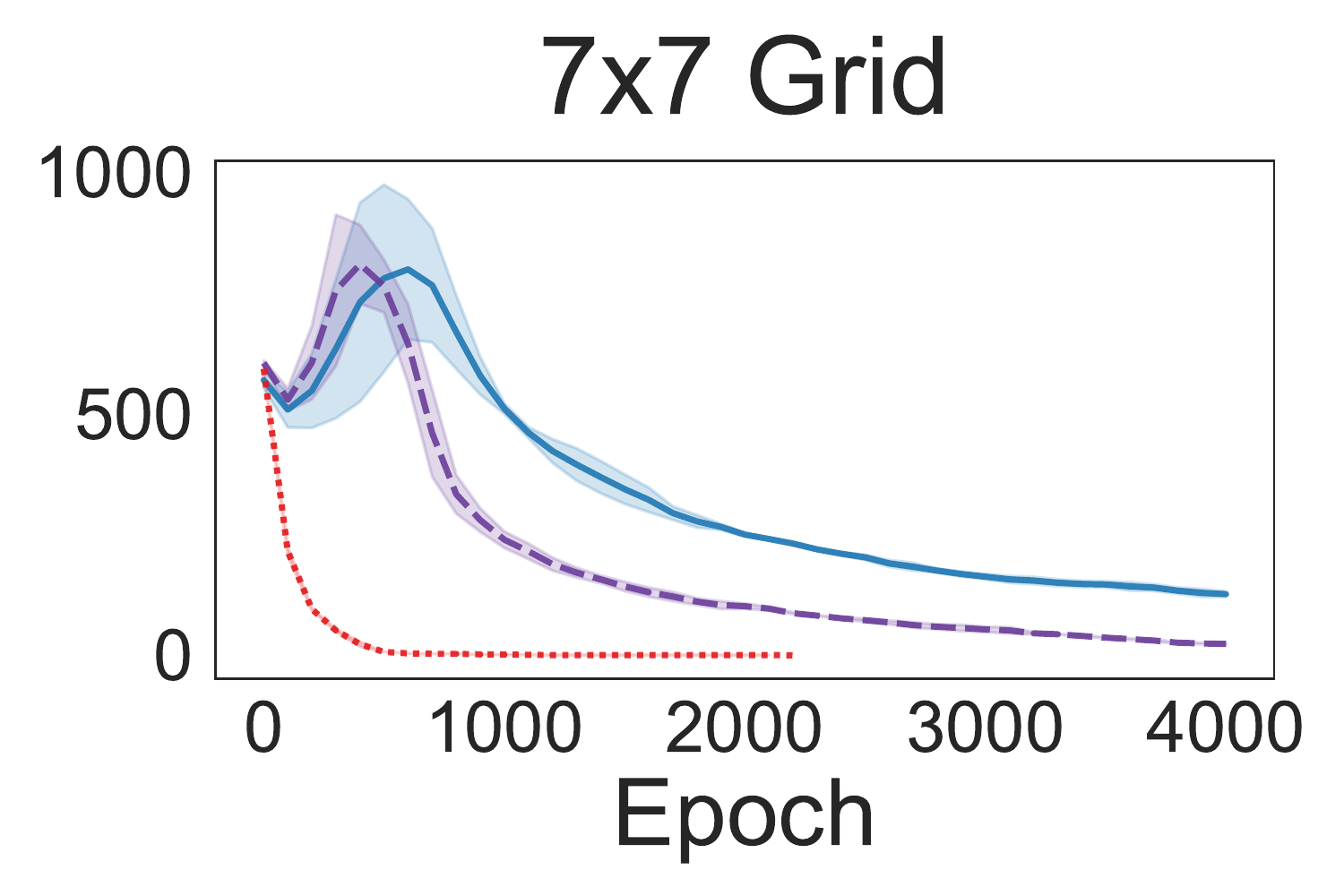}
\includegraphics[height=2.2cm]{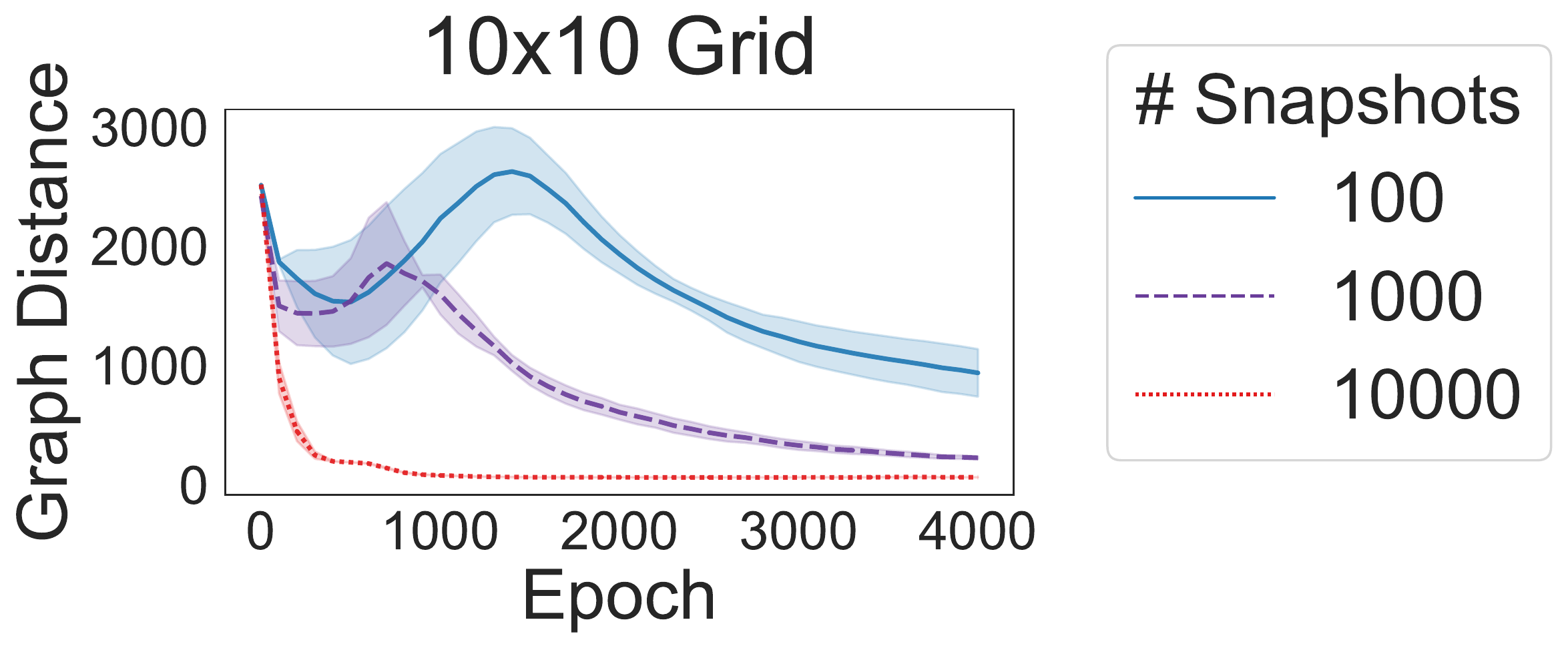}
\caption{\textbf{Exp.~2:} Influence of sample size on SIS-dynamics. $x$-axis: Epoch number. $y$-axis: Graph distance between ground truth graph and the state of \texttt{GINA}. Clearly, a higher sample size yields a better performance. CIs are based on 3 independent training runs.}
 \label{fig:exp2}
\end{figure}

We tested the effect of the number of snapshots on the graph loss using SIS-dynamics. Unsurprisingly, a larger sample size yields a better reconstructing accuracy and the training procedure converges significantly faster. For this experiment, we used a $L \times L$ grid graph with $L \in \{5,7,10\}$ and compared results based on $10^2, 10 ^3, 10^4$ snapshots.

\paragraph{Experiment 3: Comparisons with baselines\label{par:env}}

%
% Results table
%
\renewcommand{\arraystretch}{1.0}
\begin{table}
  \caption{\textbf{Exp.~3}: [Smaller is better] Results of different graph inference methods.}
  \label{sample-table}
  \centering
  \begin{tabular}{lllllrlllr}
    \toprule
    & &  \multicolumn{4}{c}{Graph loss}   & \multicolumn{4}{c}{Runtime (s)}   \\
    \cmidrule(r){3-6}
    \cmidrule(r){7-10}
    Model     & Graph  & \parbox{1.6cm}{$\texttt{GINA}$ \\ {\small (our method)}}      & Corr  & MI  & ParCorr  
    & \texttt{GINA}   & Corr  & MI  & ParCorr  \\
    \midrule   
    
    \multirow{4}{*}{SIS} 
       & ER  & \textbf{0} & \textbf{0} & \textbf{0} & \textbf{0} & $249$ & $<1$ & $<1$ & $13$    \\
        & Geom  & 64 & 78 & 84 & \textbf{40} & $14774$  & $1$ & $36$ & $12550 $    \\
        & Grid  & 12 & \textbf{0} & \textbf{0} & \textbf{0} & $7933$   & $<1 $ & $9 $ & $1843 $    \\
        & WS  & \textbf{0} & \textbf{0} &\textbf{0} & \textbf{0} & $755$  & $ <1$ & $2 $ & $125 $    \\
    \midrule 
    
    \multirow{4}{*}{Majority-Flip} 
        & ER  & \textbf{12} & 72 & 28 & 76 & $494$   & $<1 $ & $<1 $ & $11 $    \\
        & Geom  & \textbf{247} & 1130 & 616 & 1168 & $15347$  & $1 $ & $35 $ & $12655 $    \\
        & Grid  & 511 & 304 & \textbf{88} & 304 & $8184$   & $<1 $ & $8 $ & $1834 $    \\
        & WS  & \textbf{0} & 226 & 22 & 226 & $980$  & $<1 $ & $1 $ & $122 $    \\
     \midrule 
 
    \multirow{4}{*}{InvVoter} 
        & ER  & \textbf{0} & 88 & 12 & 88  & $198 $  & $ <1$ & $<1 $ & $12 $    \\
        & Geom  & \textbf{0} & 1692 & 116 & 1692   & $1682 $   & $ 1$ & $ 40$ & $12872 $    \\
        & Grid  & 502 & 360 & \textbf{90} & 360    & $ 3211$   & $ <1$ & $10 $ & $ 1835$    \\
        & WS  & \textbf{0} & 226 & 4 & 226   & $269 $   & $ <1$ & $2 $ & $124 $    \\
        
    \midrule 
    
    \multirow{4}{*}{RPS} 
        & ER  & \textbf{0} & 2 & \textbf{0}  & \textbf{0}     & $234 $   & $ <1$ & $<1 $ & $12 $    \\
        & Geom  & 151 & 164 & 172 & \textbf{2}   & $ 15382$   & $ 1$ & $ 41$ & $ 12791$    \\
        & Grid  & 75 & \textbf{0}  & \textbf{0}  & \textbf{0}    & $ 8232$  & $ <1$ & $10 $ & $1835 $    \\
        & WS  & \textbf{0} & 10 & 10 & \textbf{0}   & $ 389$   & $<1 $ & $2 $ & $124 $    \\
        
    \midrule 

    \multirow{4}{*}{Forest Fire} 
        & ER  & 25 & 6 & \textbf{2} & 28    & $3326 $   & $ <1$ & $<1 $ & $12 $    \\
        & Geom  & \textbf{36} & 852 & 300 & 906   & $  15343$   & $ 1$ & $38$ & $12481 $    \\
        & Grid  & 25 & 2  & \textbf{0}  & 2   & $ 8225$   & $ <1$ & $9 $ & $1867 $    \\
        & WS  & \textbf{0} & 2  & \textbf{0} & 8   & $ 440$   & $<1 $ & $2 $ & $120 $    \\
    \midrule 

    \multirow{4}{*}{CML} 
        & ER  & \textbf{0} & \textbf{0} & 2 & \textbf{0}    & $226 $   & $<1 $ & $<1 $ & $ 11$    \\
        & Geom  & 2 & \textbf{0} & 102 & \textbf{0}  & $8200 $  & $ 1$ & $42 $ & $ 12499$    \\
        & Grid  & 8 &\textbf{0}  & \textbf{0}  & \textbf{0}    & $8621 $   & $1 $ & $10 $ & $ 1815$    \\
        & WS  & \textbf{0} & \textbf{0}  & 4 & \textbf{0}   & $ 332$   & $ <1$ & $ 2$ & $132 $    \\
        
    \bottomrule
  \end{tabular}
\end{table}

Next, we compare \texttt{GINA} with statistical baselines.
We use a mini-batch size of $100$. We start with a sharpness parameter $v=5.0$ and increase $v$ after $50$ epochs by one. 
We train maximally for $10^4$ epochs but stop early if the underlying (binarized) graph does not change for $500$ epochs (measured each $50$ epochs). Moreover, we use Pytorch's Adam optimizer with an initial learning rate of $10^{-4}$.

%TODO: 
%- time
%- unfair
%- baseline on binaryzation
%- AIDD

%\paragraph{Ground truth graphs}
To generate ground truth graphs we use Erd\H{o}s-Renyi (ER) ($N=25$, $|E| = 44$), Geometric (Geom) ($N=200$, $|E| = 846$), and Watts–Strogatz (WS) ($N=50$, $|E| = 113$). Moreover, we use a 2$D$-grid graph with $10 \times 10$ nodes ($N=100$, $|E| = 180$). We use $50$ thousand samples. Graphs were generated the using \emph{networkX} package \cite{hagberg2008exploring} (cf.~Appendix~\ref{app:graphs} for details).

%\paragraph{Baselines}
As statistical baselines, we use Python package \emph{netrd} \cite{hartle2020network}. Specifically, we use the correlation (Corr), mutual information (MI), and partial correlation (ParCorr) methods. The baselines only return weighted matrices. Hence, they need to be binarized using a threshold. To find the optimal threshold, we provide them with the number of edges of the ground truth graph. Notably, especially in sparse graphs, this leads to unfair advantage and renders the results not directly comparable. Furthermore, \emph{netrd} only accepts binary or real-valued node-states. This is a problem for the categorical models RPS and FF. As a solution, we simply map the three node-states to real values ($1$,$2$,$3$), breaking statistical convention. Interestingly, the baselines handle this well and, in most cases, identify the ground truth graph nevertheless.       

We also compared our method to the recently proposed AIDD \cite{zhang2021automated} (Results not shown).
However, 
%while \texttt{GINA} can work on snapshots collected at arbitrary time points, 
AIDD requires \emph{consecutive} observations, making it sensitive to the time resolution of the observations. 
%For example, on a $5 \times 5$ grid graph with RPS dynamics, we find that AIDD works well when all timesteps are observable (final graph distance: 22, using 7000 samples). However, when assuming that only every tenth or 100th timestep can be observed, reconstruction quality deteriorates significantly (graph distances: 43 (every 10th), 194 (every 100th)). 
Also, we find that AIDD has a significantly higher run time than \texttt{GINA} (50 thousand samples were not feasible on our machine), which is presumably due to its more complicated architecture, particularly the alternating training of the interaction graph representation and the dynamics predictor.

\paragraph{Discussion}
The results clearly show that graph inference based on independent snapshots is possible and that \texttt{GINA} is a viable alternative to statistical methods.
Compared to the baselines, \texttt{GINA} performed best most often, even though we gave the baseline methods the advantage of knowing the ground truth number of edges. 
\texttt{GINA} performed particularly well in the challenging cases where neighboring nodes do not tend to be in the same (or in similar) node-states.
\texttt{GINA} even performed acceptably in the case of CML dynamics  despite the discretization and the chaotic and deterministic nature of the process.  

%- what is gina actually finding

\section{Conclusions and future work}
We propose \texttt{GINA}, a model-free approach to infer the underlying graph structure of a dynamical system from independent observational data. 
\texttt{GINA} is based on the principle that local interactions among agents manifest themselves in specific local patterns. These patterns can be found and exploited. 
%Specifically, we aim at finding the graph that maximizes the likelihood of the given data by learning conditional probabilities of node-states w.r.t.\ their neighborhood. 
Our experiments show that the underlying hypothesis is a promising graph inference paradigm and that \texttt{GINA} efficiently solves the task.
We believe that the main challenge for future work is to find ways of inferring graphs   when the types of interaction differ largely  among all edges. Moreover, a deeper theoretical understanding of which processes produce meaningful statistical associations, not only over time but within snapshots would be desirable. 

\newpage

\begin{ack}
This work was partially funded by the DFG project MULTIMODE.
We thank Thilo Krüger for his helpful comments on the manuscript.
\end{ack}

% \section*{References}

{
\small

\bibliography{sample}

}

%%%%%%%%%%%%%%%%%%%%%%%%%%%%%%%%%%%%%%%%%%%%%%%%%%%%%%%%%%%%
\section*{Checklist}

%%% BEGIN INSTRUCTIONS %%%
The checklist follows the references.  Please
read the checklist guidelines carefully for information on how to answer these
questions.  For each question, change the default \answerTODO{} to \answerYes{},
\answerNo{}, or \answerNA{}.  You are strongly encouraged to include a {\bf
justification to your answer}, either by referencing the appropriate section of
your paper or providing a brief inline description.  For example:
\begin{itemize}
  \item Did you include the license to the code and datasets? \answerYes{...}
  \item Did you include the license to the code and datasets? \answerNo{The code and the data are proprietary.}
  \item Did you include the license to the code and datasets? \answerNA{}
\end{itemize}
Please do not modify the questions and only use the provided macros for your
answers.  Note that the Checklist section does not count towards the page
limit.  In your paper, please delete this instructions block and only keep the
Checklist section heading above along with the questions/answers below.
%%% END INSTRUCTIONS %%%

\begin{enumerate}

\item For all authors...
\begin{enumerate}
  \item Do the main claims made in the abstract and introduction accurately reflect the paper's contributions and scope?
    \answerYes{}
  \item Did you describe the limitations of your work?
    \answerYes{See paragraph \emph{Limitations} in Section \ref{sec:Gina}}
  \item Did you discuss any potential negative societal impacts of your work?
    \answerNo{}
  \item Have you read the ethics review guidelines and ensured that your paper conforms to them?
    \answerYes{}
\end{enumerate}

\item If you are including theoretical results...
\begin{enumerate}
  \item Did you state the full set of assumptions of all theoretical results?
    \answerNA{We provide no theoretical results. }
	\item Did you include complete proofs of all theoretical results?
    \answerNA{}
\end{enumerate}

\item If you ran experiments...
\begin{enumerate}
  \item Did you include the code, data, and instructions needed to reproduce the main experimental results (either in the supplemental material or as a URL)?
    \answerYes{See the GitHub repo. Reproduction might be subject to statistical noise.}
  \item Did you specify all the training details (e.g., data splits, hyperparameters, how they were chosen)?
    \answerYes{See Section \ref{sec:experiments}}
	\item Did you report error bars (e.g., with respect to the random seed after running experiments multiple times)?
    \answerNo{Results are based on a single training iteration.}
	\item Did you include the total amount of compute and the type of resources used (e.g., type of GPUs, internal cluster, or cloud provider)?
    \answerYes{We report he machine environment and runtime in paragraph \emph{Experiment 3:  Comparisons with baselines} and Table 1.}
\end{enumerate}

\item If you are using existing assets (e.g., code, data, models) or curating/releasing new assets...
\begin{enumerate}
  \item If your work uses existing assets, did you cite the creators?
    \answerNA{}{}
  \item Did you mention the license of the assets?
    \answerNA{}
  \item Did you include any new assets either in the supplemental material or as a URL?
    \answerNA{}
  \item Did you discuss whether and how consent was obtained from people whose data you're using/curating?
    \answerNA{}
  \item Did you discuss whether the data you are using/curating contains personally identifiable information or offensive content?
    \answerNA{}
\end{enumerate}

\item If you used crowdsourcing or conducted research with human subjects...
\begin{enumerate}
  \item Did you include the full text of instructions given to participants and screenshots, if applicable?
    \answerNA{}
  \item Did you describe any potential participant risks, with links to Institutional Review Board (IRB) approvals, if applicable?
    \answerNA{}
  \item Did you include the estimated hourly wage paid to participants and the total amount spent on participant compensation?
    \answerNA{}
\end{enumerate}

\end{enumerate}

%%%%%%%%%%%%%%%%%%%%%%%%%%%%%%%%%%%%%%%%%%%%%%%%%%%%%%%%%%%%

\appendix

\section{Appendix}

\subsection{Prediction layer visualization}
%
% example Output
%
\begin{figure}[t]
\centering
\includegraphics[width=0.7\linewidth]{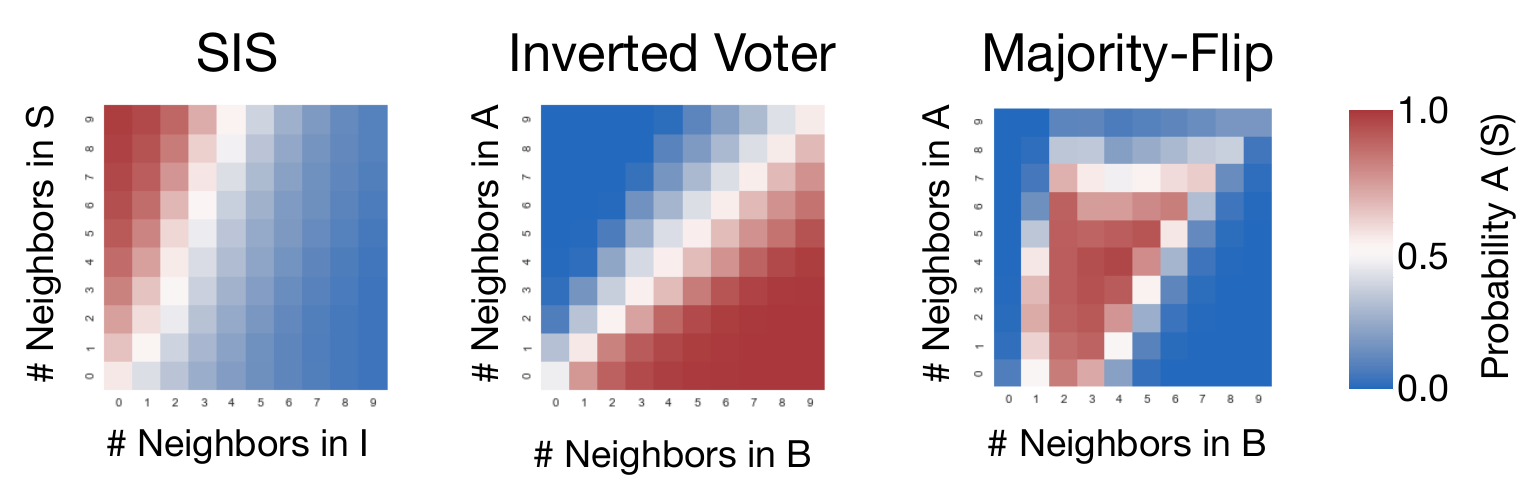}
\caption{Output of the prediction layer for a random node in the Watts–Strogatz network. The predicted probability for the node having state S in the SIS model or state A in the Inverted Voter and Majority-Flip are given.}
 \label{fig:predictOut}
\end{figure}

We can visualize the prediction layer for the 2-state models.
The input of the prediction layer of node $v_i$ is the (relaxed) 2-dimensional neighborhood counting vector $m_i$.
The probability distribution over the two node-states is fully specified by one probability.
In Fig.~\ref{fig:predictOut}, we visualize these probabilities given by the prediction layer of all possible neighborhood counting vectors (for nodes with degree $\leq 10$).
The results are given for a Watts–Strogatz graph and the same node for all three models.
We observe how the prediction layer captures a roughly linear boundary for the SIS and the Inverted Voter model.
The majority flip dynamic leads to a more intricate predictive structure.

We see that the network successfully learns to represent complicated conditional probability distributions of node-states.  

\subsection{Dynamical models \label{sec:dynmodels}}
Except for the CML, we use continuous time stochastic processes with a discrete state-space to generate snapshots.
Specifically, these models have Markov jump process or continuous-time Markov chain (CTMC) semantics. 
Moreover, each node/agent occupies one of several node-states (denoted $\mathcal{S}$) at each point in time.
Nodes change their state stochastically according to their neighborhood (more precisely, to their neighborhood counting vector).
We assume that all nodes obey the same rules/local behavior. 
We refer the reader to \cite{kiss2017mathematics,fennell2019multistate,grossmann2019reducing} for a detailed description of the CTMC construction in the context of epidemic spreading processes. 

\paragraph{SIS} Nodes are susceptible (S) or infected (I).
Infected nodes can infect their susceptible neighbors or spontaneously become susceptible again.
In other words, I-nodes become S-nodes with a fixed reaction rate $\mu$ and S-nodes become I-nods with a rate $\beta m[\text{I}]$, where $m[\text{I}]$ denotes the number of infected neighbors of the node and $\beta$ is the reaction rate parameter. 
Moreover, for all models, we add a small amount of stochastic noise $\epsilon$ to the dynamics.
The noise not only emulates measurement errors but also prevents the system from getting stuck in \emph{trap state} where no rule is applicable (e.g., all nodes are susceptible). 

In the sequel, we use the corresponding notation 
\begin{equation*}
     \text{I}\xrightarrow[]{\mu + \epsilon}  \text{S}  \phantom{XXX} 
     \text{S} \xrightarrow[]{\beta m[\text{I}]+ \epsilon}\text{I} \;.
\end{equation*}
The reaction rate refers to the exponentially distributed residence times and a higher rate is associated with a faster state transition.
When the reaction rate is zero (e.g., when no neighbors are infected), the state transition is impossible. 
  
The parameterization is $\mu = 2.0$, $\beta = 1.0$, and $\epsilon = 0.1$.
   
\paragraph{Inverted Voter}  
The Inverted Voter models two competing opinions (A and B) while nodes always tend to maximize their disagreement with their neighbors. 
\begin{equation*}
     \text{A}\xrightarrow[]{ m[\text{A}]+ \epsilon}  \text{B}  \phantom{XXX} 
     \text{B} \xrightarrow[]{ m[\text{B}]+ \epsilon}\text{A} \;.
\end{equation*}
We use $\epsilon = 0.01$.

\paragraph{Majority-Flip}  
Majority-Flip models two competing opinions while nodes tend to flip (i.e., change) their state when the majority of their neighbors follow are in the same state. A light asymmetry makes the problem solvable. 

\begin{equation*}
     \text{A}\xrightarrow[]{ \mathds{1}_{\textbf{X}} + \epsilon}  \text{B}  \phantom{XXX} 
     \text{B} \xrightarrow[]{ \mathds{1}_{\textbf{Y}} + \epsilon}\text{A} \;.
\end{equation*}
The indicator variables identifies what counts as a majority. 
They are defined as 
\begin{align*}
& \mathds{1}_{\textbf{X}} = 1 \;\; \text{iff. } \;\; \frac{m[\textbf{A}]}{m[\textbf{A}]+m[\textbf{B}]} < 0.2 \;\; \textbf{or} \;\; \frac{m[\textbf{A}]}{m[\textbf{A}]+m[\textbf{B}]} > 0.8 \\
& \mathds{1}_{\textbf{Y}} = 1 \;\; \text{iff. } \;\; \frac{m[\textbf{A}]}{m[\textbf{A}]+m[\textbf{B}]} < 0.3 \;\; \textbf{or} \;\; \frac{m[\textbf{A}]}{m[\textbf{A}]+m[\textbf{B}]} > 0.7 \;.
\end{align*}
If the indicator is not one, it is zero. 
We use $\epsilon = 0.01$.

\paragraph{Rock Paper Scissors} 
provides a simple evolutionary dynamics where three species overtake each other in a ring-like relationship. 
\begin{equation*}
     \text{R}\xrightarrow[]{ m[\text{P}] + \epsilon}  \text{P}  \phantom{XXX} 
     \text{P}\xrightarrow[]{ m[\text{S}]+ \epsilon}  \text{S}  \phantom{XXX} 
     \text{S} \xrightarrow[]{ m[\text{R}]+ \epsilon}  \text{R}  \;.
\end{equation*}
We use $\epsilon = 0.01$.

\paragraph{Forest Fire} 

Spots/nodes are either empty (E), on fire (F), or have a tree on them (F).
Trees grow with a growth rate $g$. 
Random lightning starts a fired on tree-nodes with rate $f_{\text{start}}$.
The fire on a node goes extinct with rate $f_{\text{end}}$ leaving the node empty. Finally, fire spreads to neighboring tree-nodes with 
rate $f_{\text{spread}}$.

\begin{equation*}
     \text{T}\xrightarrow[]{ f_{\text{start}} + m[\text{F}]f_{\text{spread}} + \epsilon }  \text{F}  \phantom{XXX} 
     \text{F}\xrightarrow[]{f_{\text{end} +\epsilon} }  \text{E} 
     \phantom{XXX} 
     \text{E}\xrightarrow[]{g +\epsilon }  \text{T}  \;.
\end{equation*}

The parameterization is $g = 1.0$, $f_{\text{start}} = 0.1$, $f_{\text{end}} = 2.0$,
$f_{\text{spread}} = 2.0$, and $\epsilon = 0.1$.

\paragraph{Coupled Map Lattice} 
Let $x_i$ be the value of node $v_i$ at time-step $i$.
Each node starts with a random value (uniform in $[0,1]$).
At each time step all nodes are updated based on a linear combination of the node-value and the node-values of neighboring nodes \cite{kaneko1992overview}:
\begin{equation*}
x_{i+1} = (1.0-s) f(x_i) + \frac{s}{d_i} \sum_{j \in N(i)} f(x_j) \;,
\end{equation*}
where $d_i$ is the degree of $v_i$, $N(i)$ denotes the set of (indices of) nodes adjacent to $v_i$, $s$ is the coupling strength and $f$ is the local map. 
Like \cite{zhang2019general}, we use the logistic function \cite{may2004simple}:
\begin{equation*}
f(x)= r \cdot x \cdot (1.0-x) \;.
\end{equation*}
where $r$ modulates the complexity of the dynamics.

We use $s=0.1$ and $r=3.57$.
%return r*x*(1.0-x)

% (1.0-s) * f_map(v) + s/len(neig) * np.sum([f_map(states[n_j]) for n_j in neig])

\subsection{Random interaction graphs\label{app:graphs}}
We use the Python NetworkX \cite{hagberg2008exploring} package to generate a single instance (variate) of a random graph model and test the \text{GINA} and the baselines and a large number of snapshots generated using this graph.
In particular, we use Erd\H{o}s-Renyi (ER) ($N=25$, $|E| = 44$) graph model with connection probability $0.15$:
\begin{verbatim}
    nx.erdos_renyi_graph(25, 0.15, seed=43)
\end{verbatim}
(the 43 seed was used in order to create a connected graph).
We also use 
Geometric Graph ($N=200$, $|E| = 846$):
\begin{verbatim}
    nx.random_geometric_graph(200, 0.125, seed=42)
\end{verbatim}
and a Watts–Strogatz graph ($N=50$, $|E| = 113$) where each node has 4 neighbors and the re-wiring probability is $0.15$:
\begin{verbatim}
    nx.newman_watts_strogatz_graph(50, 4, 0.15, seed=42)
\end{verbatim}
After generation, the node-ids are randomly shuffled in order to guarantee that they do not leak information about connectivity to the training model. 

\end{document}